\documentclass[journal]{IEEEtran}
\ifCLASSINFOpdf
\else
\fi
\usepackage{tabularx} 
\usepackage{arydshln}

\usepackage{algorithm}
\usepackage{algorithmic}

\usepackage{dsfont}

\usepackage{amsfonts}
\usepackage{amssymb}
\usepackage{amsmath}
\usepackage{mathdots}
\usepackage{mathtools}
\usepackage{amsthm}
\usepackage{bbm}
\usepackage{bm}
\usepackage{mathrsfs}
\usepackage{balance}

\usepackage{url}
\usepackage{enumitem}
\usepackage{multirow,tabularx}

\usepackage{booktabs}

\usepackage{graphicx}
\usepackage{subfig}
\usepackage{epstopdf}

\newtheorem{open problem}{Open Problem}

\begin{document}
%
\title{Learning Graph Embedding with Adversarial Training Methods}
%
%

	\author{Shirui Pan,
	Ruiqi Hu,
	Sai-fu Fung,
	Guodong Long,
	Jing Jiang,
	and~Chengqi Zhang,~\IEEEmembership{Senior Member,~IEEE}
	
	\IEEEcompsocitemizethanks{
		\IEEEcompsocthanksitem S. Pan is with Faculty of Information Technology, Monash University, Clayton, VIC 3800, Australia (Email: shirui.pan@monash.edu). 
		\IEEEcompsocthanksitem R. Hu, G. Long, J. Jiang, C. Zhang are with Centre for Artificial Intelligence, FEIT, University of Technology Sydney, NSW 2007, Australia (E-mail: ruiqi.hu@uts.edu.au;  jing.jiang@uts.edu.au; guodong.long@uts.edu.au; chengqi.zhang@uts.edu.au).
		\IEEEcompsocthanksitem S.F. Fung is with  Department of Applied Social Sciences, City University of Hong Kong, China.(E-mail: sffung@cityu.edu.hk).
		\IEEEcompsocthanksitem Corresponding author: Ruiqi Hu.
	}
	
	\thanks{Manuscript received April 19, 201x; revised August 26, 201x.}}

%
%

\markboth{IEEE Transactions on Cybernetics,~Vol.~xx, No.~xx, July~2019}%
{Shell \MakeLowercase{\textit{et al.}}: Bare Demo of IEEEtran.cls for IEEE Journals}
%



\maketitle

\begin{abstract}
			Graph embedding aims to transfer a graph into vectors to facilitate subsequent graph analytics tasks like link prediction and graph clustering. Most approaches on graph embedding focus on preserving the graph structure or minimizing the reconstruction errors for graph data. They have mostly overlooked the embedding distribution of the latent codes, which unfortunately may lead to inferior representation in many cases. In this paper, we present a novel adversarially regularized framework for graph embedding. By employing the graph convolutional network as an encoder, our framework embeds the topological information and node content into a vector representation, from which a graph decoder is further built to reconstruct the input graph. The adversarial training principle is applied to enforce our latent codes to match a prior Gaussian or Uniform distribution. Based on this framework, we derive two variants of adversarial models, the adversarially regularized graph autoencoder (\textbf{ARGA}) and its variational version, adversarially regularized variational graph autoencoder (\textbf{ARVGA}), to learn the graph embedding effectively. We also exploit other potential variations of ARGA and ARVGA to get a deeper understanding on our designs.  Experimental results compared among twelve algorithms for link prediction and twenty algorithms for graph clustering validate our solutions.
\end{abstract}


%
\IEEEpeerreviewmaketitle

	\section{Introduction}
	\IEEEPARstart{G}{raphs} are essential tools to capture and model  complicated relationships among data.  In a variety of graph applications, such as social networks, citation networks, protein-protein interaction networks,   graph data analysis plays an important role in various data mining tasks including  classification \cite{kipf2016semi}, 
	 clustering \cite{wang2017mgae},  recommendation \cite{xiong2018social,cai2017comprehensive,shi2018heterogeneous}, and graph classification \cite{Pantkde15,Pantkde16}. However, the high computational complexity, low parallelizability, and inapplicability of machine learning methods to graph data have made these graph analytic tasks profoundly challenging \cite{cui2017survey,shi2017survey}. \textit{Graph embedding} has recently emerged as a general approach to these problems.
	
	Graph embedding transfers graph data into a low dimensional, compact, and continuous feature space. The fundamental idea is to preserve the topological structure, vertex content, and other side information \cite{zhang2017network, wang2019heterogeneous}. This new learning paradigm has shifted the tasks of seeking complex models for classification, clustering, and link prediction \cite{cao2016link} to learning a compact and informative representation for the graph data, so that many graph mining tasks can be easily performed by employing simple traditional models (e.g.,  a linear SVM for the classification task). This merit has motivated many studies in this area \cite{cai2017comprehensive,goyal2017graph}.

	Graph embedding algorithms can be classified into three categories:  probabilistic models, matrix factorization-based algorithms, and deep learning-based algorithms. Probabilistic models like DeepWalk \cite{perozzi2014deepwalk}, node2vec \cite{grover2016node2vec} and LINE \cite{Tang2015} attempt to learn graph embedding by extracting different patterns from the graph.  The captured patterns or walks include global structural equivalence, local neighborhood connectivities, and other various order proximities. Compared with classical methods such as Spectral Clustering \cite{tang2011leveraging}, these graph embedding algorithms perform more effectively and are scalable to large graphs.

	Matrix factorization-based algorithms, such as GraRep \cite{cao2015grarep}, HOPE \cite{ou2016asymmetric}, M-NMF \cite{wang2017community} pre-process the graph structure into an adjacency matrix and obtain the embedding by factorizing the adjacency matrix.  It has been recently shown that many probabilistic algorithms including DeepWalk \cite{perozzi2014deepwalk}, LINE \cite{Tang2015}, node2vec \cite{grover2016node2vec}, are equivalent to matrix factorization approaches \cite{qiu2017network}, and Qiu et al. propose a unified matrix factorization approach NetMF \cite{qiu2017network} for graph embedding. Deep learning approaches, especially autoencoder-based methods,  are also  studied for graph embedding (a most up-to-date survey on graph neural networks can be found here \cite{Wu-GNNsurvey-19}). SDNE \cite{wang2016structural} and DNGR \cite{cao2016deep} employ deep autoencoders to preserve the graph proximities and model the positive pointwise mutual information (PPMI). The MGAE algorithm utilizes a marginalized single layer autoencoder to learn representation for graph clustering \cite{wang2017mgae}. The DNE-SBP model is proposed for signed network embedding with a stacked auto-encoder framework \cite{Shen-TCYB18}.

	The approaches above are typically unregularized approaches which mainly focus on preserving the structure relationship (probabilistic approaches) or minimizing the reconstruction error (matrix factorization or deep learning methods). They have mostly ignored the latent data distribution of the  representation. In practice, unregularized embedding approaches often learn a degenerate \textit{identity} mapping where the latent code space is free of any structure \cite{makhzani2015adversarial}, and can easily result in poor representation in dealing with real-world sparse and noisy graph data. One standard way to handle this problem is to introduce some regularization to the latent codes and enforce them to follow some prior data distribution \cite{makhzani2015adversarial}. Recently generative adversarial based frameworks  \cite{donahue2016adversarial,zhao2016energy,dumoulin2016adversarially,radford2015unsupervised} have also been developed for learning robust latent representation. However, none of these frameworks is specifically for graph data, where both topological structure and content information are required to be represented into a latent space.

    In this paper, we propose a novel adversarially regularized algorithm with two variants, \textit{adversarially regularized graph autoencoder} (ARGA) and its variational version, \textit{adversarially regularized variational graph autoencoder} (ARVGA), for graph embedding. The theme of our framework is to not only minimize the reconstruction errors of the topological structure but also to enforce the learned latent embedding to match a prior distribution.  By exploiting both graph structure and node content with a graph convolutional network, our algorithms encode the graph data in the latent space. With a decoder aiming at reconstructing the topological graph information,  we further incorporate an adversarial training scheme to regularize the latent codes to learn a robust graph representation. The adversarial training module aims to discriminate if the latent codes are from a real prior distribution or the graph encoder. The graph encoder learning and adversarial regularization learning are jointly optimized in a unified framework so that each can be beneficial to the other and finally lead to a better graph embedding. To get further insight into the influence of prior distribution, we have varied it with the Gaussian distribution and Uniform distribution for all models and tasks. Moreover, we have examined the different ways to construct the graph decoders as well as the target of the reconstructions. By doing so, we have obtained a comprehensive view of the most influential factor of the adversarially regularized graph autoencoder models for different tasks. The experimental results on three benchmark graph datasets demonstrate the superb performance of our algorithms on two unsupervised graph analytic tasks, namely link prediction and node clustering. Our contributions can be summarized below:

	
	\begin{itemize}
		
        \item We propose a novel adversarially regularized framework for graph embedding, which represents topological structure and node content in a continuous vector space. Our framework learns the embedding to minimize the reconstruction error while enforcing the latent codes to match a prior distribution. 
        
        \item We develop two variants of adversarial approaches, \textit{adversarially regularized graph autoencoder} (ARGA) and \textit{adversarially regularized variational graph autoencoder} (ARVGA) to learn the graph embedding. 
        
        \item We have examined different prior distributions, the ways to construct decoders, and the targets of the reconstructions to point out the influence of the factors of the adversarially regularized graph autoencoder models on various tasks.
        
        \item  Experiments on benchmark graph datasets demonstrate that our graph embedding approaches outperform the others on different unsupervised tasks.
	\end{itemize}
	
	The paper is structured as follows. Section \ref{sec:related} reviews the related work. Section \ref{sec:problem} outlines the problem definition and our overall framework. Section \ref{sec:algorithm} presents the proposed algorithm and Section \ref{sec:experiments} describes the experimental results. We conclude the paper in Section \ref{sec:conclusion}.
	%
	%
	%
	%
	\section{Related Work} \label{sec:related}
	
	\subsection{Graph Embedding Models} 
	Graph embedding, also known as network embedding \cite{cai2017comprehensive} or network representation learning \cite{zhang2017network}, transfers a  graph into vectors. 
	From the perspective of information exploration, graph embedding algorithms can be separated into two groups: topological network embedding approaches and content enhanced network embedding methods. 
	
	\vspace{.1cm}
	\noindent\textbf{Topological network embedding approaches} 
    Topological network embedding approaches assume that there is only topological structure information available, and the learning objective is to preserve the topological information maximumly \cite{zhu2018high, gui2017embedding}. Inspired by the word embedding approach \cite{mikolov2013efficient}, Perozzi et al. propose a DeepWalk model to learn the node embedding from a collection of random walks \cite{perozzi2014deepwalk}. Since then, many probabilistic models have been developed. Specifically, Grover et al. propose a biased random walks approach, node2vec \cite{grover2016node2vec}, which employs both breadth-first sampling (BFS) and Depth-first sampling (DFS) strategies to generate random walk sequences for network embedding. Tang et al. propose a LINE algorithm \cite{Tang2015} to handle large-scale information networks while preserving both first-order and second-order proximity. Other random walk variants include hierarchical representation learning approach (HARP) \cite{chen2017harp}, and discriminative deep random walk (DDRW) \cite{li2016discriminative}, and Walklets \cite{perozzi2016walklets}.	
	
	Because a graph can be mathematically represented as an adjacency matrix, many matrix factorization approaches are proposed to learn the latent representation for a graph. GraRep \cite{cao2015grarep} integrates the global topological information of the graph into the learning process to represent each node into a low dimensional space; HOPE \cite{ou2016asymmetric} preserves the asymmetric transitivity by approximating high-order proximity for a better performance on capturing topological information of graphs and reconstructing from partially observed graphs; DNE \cite{Shen-ijcai-18}  aims to learn discrete embedding which reduces the storage and computational cost. Recently deep learning models have been exploited to learn the graph embedding. These algorithms preserve the first and second order of proximities \cite{wang2016structural}, or reconstruct the positive pointwise mutual information (PPMI) \cite{cao2016deep} via different variants of autoencoders. 
	
	
	\vspace{.1cm}
	\noindent\textbf{Content enhanced network embedding methods} 
    Content enhanced embedding methods assume node content information is available and exploit both topological information and content features simultaneously. TADW \cite{yang2015network} proved that DeepWalk can be interpreted as a  factorization approach and proposed an extension to DeepWalk to explore node features. TriDNR \cite{pan2016tri} captures structure, node content, and label information via a tri-party neural network architecture. UPP-SNE \cite{zhang2017user} employs an approximated kernel mapping scheme to exploit user profile features to enhance the embedding learning of users in social networks. SNE \cite{Liao2018} learns a neural network model to capture both structural proximity and attribute proximity for attributed social networks. DANE \cite{li2017attributed} deals with the dynamic environment with an incremental matrix factorization approach, and LANE \cite{huang2017label} incorporates label information into the optimization process to learn a better embedding. Recently,  BANE \cite{ICDM-18-Yang} is proposed to learn binarized embedding for an attributed graph which has the potential to increase the efficiency for latter graph analytic tasks. 
    
    Although these algorithms are well-designed for graph-structured data, they have largely ignored the embedding distribution, which may result in poor representation in real-graph data. In this paper, we explore adversarial training approaches to address this issue.

	\subsection{Adversarial Models}
    Our method is motivated by the generative adversarial network (GAN) \cite{goodfellow2014generative}. GAN plays an adversarial game with two linked models: the generator $\mathcal{G}$ and the discriminator $\mathcal{D}$. The discriminator discriminates if an input sample comes from the prior data distribution or from the generator we built. Simultaneously, the generator is trained to generate the samples to convince the discriminator that the generated samples come from the prior data distribution. Typically, the training process is split into two steps: (1) Train the discriminator $\mathcal{D}$ for iterations to distinguish the samples from the expected data distribution from the samples generated via the generator. Then (2) train the generator to confuse the discriminator with its generated data. However, the original GAN does not fit the unsupervised data encoding, as the absence of the precise structure for inference. To implement the adversarial structure in learning data embedding, existing works like BiGAN\cite{donahue2016adversarial}, EBGAN\cite{zhao2016energy} and ALI\cite{dumoulin2016adversarially} arrive at extending the original adversarial framework with external structures for the inference, which have achieved non-negligible performance in applications, such as document retrieval\cite{glover2016modeling} and image classification\cite{donahue2016adversarial}. Other solutions manage to generate the embedding from the discriminator or generator for semi-supervised and supervised tasks via reconstructed layers. For example, DCGAN\cite{radford2015unsupervised} bridges the gap between convolutional networks and generative adversarial networks with particular architectural constraints for unsupervised learning; and ANE\cite{dai2017adversarial} combines a structure-preserving component and an adversarial learning scheme to learn a robust embedding. 
	
     Makhzani et al. proposed an adversarial autoencoder (AAE) to learn the latent embedding by merging the adversarial mechanism into the autoencoder \cite{makhzani2015adversarial}. However, AAE is designed for general data rather than graph data. Recently there are some studies on applying the adversarial mechanism to graphs such as AIDW \cite{dai2018adversarial} and NetRA \cite{yu2018learning}. However, their approach can only exploit the topological information \cite{dai2017adversarial,wang2017graphgan, yu2018learning}. In contrast, our algorithm is more flexible and can handle both topological and content information for graph data. Furthermore, these models, such as NetRA, can only reconstruct the graph structure, while ARGA\_AX reconstructs both topological structure and node characteristics, smoothly persevering the integrity of the given graph through entire encoding and decoding processing.  Most recently, Ding et al. proposed a GraphSGAN \cite{ding2018semi} for semi-supervised node classification with the GAN principle, and Hu et al. proposed the HeGAN \cite{hu2019adver} for heterogeneous information network embedding. 
	 
	
	
    Though many adversarial models have achieved impressive success in computer vision,  they cannot effectively and directly handle the graph-structured data. With some preliminary study in \cite{pan2018adversarially},  we try to thoroughly exploit the graph convolutional models with different adversarial models to learn a robust graph embedding in this paper.
    
    In particular, we have proposed four new algorithms to handle networks with limited labeled data. These algorithms aim to reconstruct different content in a network, including topological structure only or both the topological structure and node content, by using general graph encoder or variational graph encoder as a building block. We also conducted more extensive experiments to validate the proposed algorithms with a wide range of metrics including NMI, ACC, F1, Precision, ARI and Recall. 
	
	\subsection {Graph Convolutional Nets based Models}
    Graph convolutional networks (GCN) \cite{kipf2016semi} is a semi-supervised framework based on a variant of convolutional neural networks, which attempt to operate the graphs directly. Specifically, the GCN represents the graph structure and the interrelationship between node and feature with an adjacent matrix $\mathbf{A}$ and node-feature matrix $\mathbf{X}$. Hence, GCN can directly embed the graph structure with a spectral convolutional function $f(\mathbf{X}, \mathbf{A})$ for each layer and train the model on a supervised target for all labelled nodes. Because of the spectral function $f(\bullet)$ on the adjacent matrix $\mathbf{A}$ of the graph, the model can distribute the gradient from the supervised cost and learn the embedding of both the labelled and unlabelled nodes.  Although GCN is powerful on graph-structured data sets for semi-supervised tasks like node classification, variational graph autoencoder VGAE \cite{kipf2016variational} extends it into unsupervised scenarios. Specifically, VGAE integrates the GCN into the variational autoencoder framework \cite{kingma2013auto} by framing the encoder with graph convolutional layers and remodeling the decoder with a link prediction layer. Taking advantage of GCN layers, VGAE can naturally leverage the information of node features, which expressively muscle the predictive performance. Recently GCN is used to learn the binary codes for improving the efficiency of information retrieval \cite{Zhou-TCYB18}. 
	\begin{figure*}
		\centering
		\includegraphics[ width=.85\linewidth]{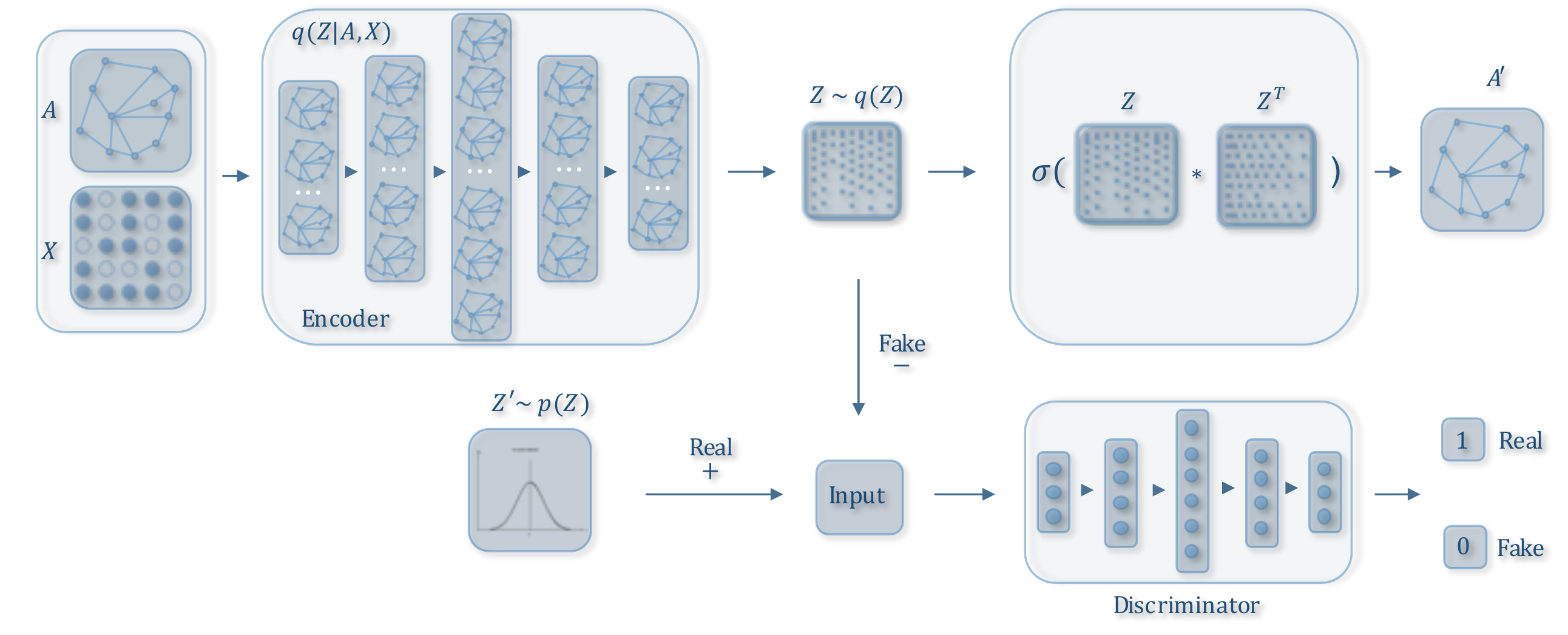}
		\caption{The architecture of the adversarially regularized graph autoencoder (\textbf{ARGA}). The upper tier is a graph convolutional autoencoder that reconstructs a graph $\mathbf{A}$ from an embedding  $\mathbf{Z}$ which is generated by the encoder which exploits graph structure $\mathbf{A}$ and the node content matrix $\mathbf{X}$. The lower tier is an adversarial network trained to discriminate if a sample is generated from the embedding or from a prior distribution.  The adversarially regularized variational graph autoencoder (ARVGA) is similar to ARGA except that it employs a \textit{variational} graph autoencoder in the upper tier (See Algorithm 1 for details). 
		} \label{fig:construction}
	\end{figure*}
	
	\section{Problem Definition and Framework}\label{sec:problem}
	A graph is represented as $\mathbf{G} = \{\mathbf{V}, \mathbf{E}, \mathbf{X}\}$, where $\mathbf{V} = \{\mathbf{v}_i\}_i = 1, \cdots, n$ is constitutive of a set of nodes in a graph and $\mathbf{e}_{i,j}=<\mathbf{v}_{i},\mathbf{v}_{j}>  \in \mathbf{E}$ represents a linkage coding the citation edge between the papers (nodes). The topological structure of graph $\mathbf{G}$ can be represented by an adjacency matrix $\mathbf{A}$, where $\mathbf{A}_{i,j} = 1$ if $\mathbf{e}_{i,j} \in \mathbf{E}$, otherwise $\mathbf{A}_{i,j} = 0$. $\mathbf{x}_i \in \mathbf{X}$ encodes the textual content features associated with each node $\mathbf{v}_i$. 
	
	Given a graph $\mathbf{G}$, our purpose is to map the nodes $\mathbf{v}_{i} \in \mathbf{V}$ to  low-dimensional vectors $\mathbf{z}_i \in \mathbbm{R}^d$ with the formal format as follows: $f: \mathbf{(A, X)} \rightarrowtail \mathbf{Z}$, where $\mathbf{z}_i^{\top}$ is the $i$-th row of the matrix $\mathbf{Z} \in \mathbbm{R}^{n \times d}$. $n$ is the number of nodes and $d$ is the dimension of embedding. We take $\mathbf{Z}$ as the embedding matrix and the embeddings should well preserve the topological structure $\mathbf{A}$ as well as content information $\mathbf{X}$.

	\subsection{Overall Framework}
	The objective is to learn a robust embedding for a given graph $\mathbf{G} = \{\mathbf{V}, \mathbf{E}, \mathbf{X}\}$. To this end, we leverage an adversarial architecture with a graph autoencoder to directly process the entire graph and learn a robust embedding. Figure \ref{fig:construction} demonstrates the workflow of ARGA which consists of two modules: the graph autoencoder and the adversarial network.
	\begin{itemize}
		\item \textbf{Graph convolutional autoencoder}. The autoencoder takes in the structure of graph $\mathbf{A}$ and the node content $\mathbf{X}$ as inputs to learn a latent representation $\mathbf{Z}$, and then reconstructs the graph structure $\mathbf{A}$ from $\mathbf Z$. We will further explore other variants of graph autoencoder in Section \ref{sec:variation}.
		\item \textbf{Adversarial regularization.} The adversarial network forces the latent codes to match a prior distribution by an adversarial training module, which discriminates whether the current latent code $\mathbf z_i \in \mathbf{Z}$ comes from the encoder or from the prior distribution. 
	\end{itemize}
	
	\section{Proposed Algorithm}\label{sec:algorithm}
	\subsection{Graph Convolutional Autoencoder}
	Our graph convolutional autoencoder aims to embed a graph $\mathbf{G} = \{\mathbf{V}, \mathbf{E}, \mathbf{X}\}$ in a low-dimensional space. Two fundamental questions arise (1) how to simultaneously integrate graph structure $\mathbf{A}$ and content feature $\mathbf{X}$ in an encoder, and (2)  what sort of information should be reconstructed via a decoder? 
	
	\vspace{.1cm}
	\noindent \textbf{Graph Convolutional Encoder Model  $\mathcal{G}(\mathbf X, \mathbf A)$. \ } 
	To represent both graph structure $\mathbf{A}$ and node content $\mathbf{X}$ in a unified framework, we  develop a variant of the graph convolutional network (GCN) \cite{kipf2016semi} as a graph encoder. GCN introduces the \textit{convolutional} operation to graph-data from the spectral area, and leverages a spectral convolutional function $f(\mathbf{Z}^{(l)},\mathbf{A} |\mathbf{W}^{(l)})$ to build a layer-wise transformation: 
	\begin{equation}
	\mathbf{Z}^{(l+1)} = f(\mathbf{Z}^{(l)},\mathbf{A} |\mathbf{W}^{(l)})
	\label{eq:encoder}
	\end{equation}
	Here, $\mathbf{Z}^l $ and $\mathbf{Z}^{(l+1)}$ are the input and output of the convolution respectively. We set $\mathbf{Z}^0 = \mathbf{X} \in \mathbbm{R}^{n\times m}$  ($n$ indicates the number of nodes and $m$ indicates the number of features) for our problem.  We need to learn a filter parameter matrix $\mathbf{W}^{(l)} $ in the neural network, and if the spectral convolution function 
	is well defined, we can efficiently construct arbitrary deep convolutional neural networks.
	
	Each layer of our graph convolutional network can be expressed with the the spectral convolution function $f(\mathbf{Z}^{(l)},\mathbf{A} |\mathbf{W}^{(l)})$ as follows:
	\begin{equation}
	\label{eq:convolutional}
	f(\mathbf{Z}^{(l)},\mathbf{A} |\mathbf{W}^{(l)})=   \phi(\widetilde{\mathbf{D}}^{-\frac{1}{2}}\widetilde{\mathbf{A}}\widetilde{\mathbf{D}}^{-\frac{1}{2}}\mathbf{Z}^{(l)}\mathbf{W}^{(l)}),
	\end{equation}
	where $\widetilde{\mathbf{D}}_{ii} = \sum_j\widetilde{\mathbf{A}}_{ij}$ and $\widetilde{\mathbf{A}} = \mathbf{A}+ \mathbf{I}$. $\mathbf{I}$ is the identity matrix of ${\mathbf{A}}$ and $\phi$ is an activation function such as $\text{sigmoid}(t)=\frac{1}{1+e^t}$ or $\text{Relu}(t) = \max(0,t)$. 
	Overall, the graph encoder $\mathcal{G}(\mathbf X, \mathbf A)$ is constructed with a two-layer GCN. In our paper, we develop two variants of the encoder, e.g., Graph Encoder and Variational Graph Encoder.
	
	The \noindent\textit{ Graph Encoder} is constructed as follows: 
	\begin{eqnarray}
	\mathbf{Z}^{(1)}= f_{\text{Relu}}(\mathbf X,\mathbf{A}|  \mathbf W^{(0)}); \label{eq:z1}\\
	\mathbf Z^{(2)} = f_{\text{linear}}(\mathbf{Z}^{(1)},\mathbf{A} | \mathbf W^{(1)}). \label{eq:latent}
	\end{eqnarray}
	$\text{Relu}(\cdot)$ and linear activation  functions are used for the first and second layers. Our graph convolutional encoder  $\mathcal{G}(\mathbf Z, \mathbf A)=q(\mathbf{Z}|\mathbf{X}, \mathbf{A})$  encodes both graph structure and node content into a  representation $\mathbf Z= q(\mathbf{Z}|\mathbf{X}, \mathbf{A})=\mathbf{Z}^{(2)}$.
	
	\vspace{.1cm}
	A \noindent\textit{Variational Graph Encoder}  is defined by an inference model:
	\begin{eqnarray}
	q(\mathbf{Z}|\mathbf{X}, \mathbf{A}) = \prod^{n}_{i=1}q(\mathbf{z_i}|\mathbf{X},\mathbf{A}),\\
	q(\mathbf{z_i}|\mathbf{X},\mathbf{A}) = \mathcal{N}(\mathbf{z}_i|  \bm{{\mu}}_i, \text{diag}(\bm \sigma^2))
	\end{eqnarray}
	
	Here, $ \bm \mu= \mathbf Z^{(2)}$ is the matrix of mean vectors $\bm z_i$
	; similarly $\text{log} \bm \sigma = f_{\text{linear}}(\mathbf{Z}^{(1)},\mathbf{A} | \mathbf W'^{(1)})$ which shares the weights $\mathbf W^{(0)}$ with $ \bm \mu$ in the first layer in Eq. (\ref{eq:z1}).
	
	\vspace{.1cm}
	\noindent \textbf{Decoder Model. \ } Our decoder model is used to reconstruct the graph data. We can reconstruct either the graph structure $\mathbf A$, content information $\mathbf X$, or both. In the basic version of our model (ARGA), we propose to reconstruct graph structure $\mathbf A$, which  provides more flexibility in the sense that our algorithm will still function properly even if there is no content information $\mathbf X$ available (e.g., $\mathbf X = \mathbf I$). We will provide several variants of decoder model in Section \ref{sec:variation}. Here the ARGA decoder $p(\hat{\mathbf{A}}|\mathbf{Z})$ predicts whether there is a link between two nodes. More specifically, we train a link prediction layer based on the graph embedding:
	\begin{eqnarray}
	p(\hat{\mathbf{A}}|\mathbf{Z}) = \prod^n_{i=1} \prod^n_{j=1} p(\hat{\mathbf{A}}_{ij}| \mathbf{z}_i,\mathbf{z}_j);\\
	p(\hat{\mathbf{A}}_{ij} = 1 |\mathbf{z}_i, \mathbf{z}_j) = \text{sigmoid}(\mathbf{z}^{\top}_i, \mathbf{z}_j),
	\end{eqnarray}
	here the prediction $\hat{\mathbf{A}}$  should be close to the ground truth $\mathbf A$.

	\vspace{.1cm}
	\noindent \textbf{Graph Autoencoder Model. \ } 
	The embedding $\mathbf{Z}$ and the reconstructed graph $\hat{\mathbf{A}}$ can be presented as follows:
	\begin{equation}
	\hat{\mathbf{A}} =\text{sigmoid}(\mathbf{Z}\mathbf{Z}^{\top}),\ \text{here} \ \mathbf{Z}  = q(\mathbf{Z}|\mathbf{X}, \mathbf{A}) 
	\end{equation}

	\noindent \textbf{Optimization. \ }  For the graph encoder, we minimize the reconstruction error of the graph data by:
	\begin{equation}
	\mathcal{L}_{0} = \mathbbm{E}_{q(\mathbf{Z|(X,A)})}[\text{log}~ p(\mathbf{A}|\mathbf{Z})] \label{eq:autoencoder_obj0}
	\end{equation}
	For the variational graph encoder, we optimize the variational lower bound as follows:
	\begin{equation}
	\mathcal{L}_{1} = \mathbbm{E}_{q(\mathbf{Z|(X,A)})}[\text{log}~ p(\mathbf{A}|\mathbf{Z})] - \mathbf{KL}[q(\mathbf{Z|\mathbf{X}},\mathbf{A}) \parallel p(\mathbf{Z})] \label{eq:autoencoder_obj}
	\end{equation}
	where $\mathbf{KL}[q(\bullet)||p(\bullet)]$ is the Kullback-Leibler divergence between $q(\bullet)$ and $p(\bullet)$. $p(\bullet)$ is a prior distribution which can be a uniform distribution or a Gaussian distribution $p(\mathbf{Z}) = \prod_i p(\mathbf{z}_i) = \prod_i\mathcal{N}(\mathbf{z}_i|0,\mathbf{I}) $ in practice. 

	\subsection{Adversarial Model $\mathcal{D}(\mathbf Z)$}
	The fundamental idea of our  model is to enforce latent representation $\mathbf Z$ to match a prior distribution, which is achieved by an adversarial training model. The adversarial model is built on a standard multi-layer perceptron (MLP) where the output layer only has one dimension with a sigmoid function. The  adversarial model acts as a discriminator to distinguish whether a latent code is from the prior $p_z$ (positive) or graph encoder $\mathcal{G}(\mathbf{X, A})$ (negative). By minimizing the  cross-entropy cost for training the binary classifier, the embedding will finally be regularized and  improved during the training process. The cost can be computed as follows:
	\begin{equation}
	-\frac{1}{2}\mathbbm{E}_{\mathbf{z}\sim p_{z}}\text{log}  \mathcal{D}(\mathbf{Z})-\frac{1}{2}\mathbbm{E}_{\mathbf{X}}\text{log}(1-\mathcal{D}(\mathcal{G}(\mathbf{X, A}))),
	\end{equation}

	In our paper, we have examined both Gaussian distribution and Uniform distribution as $p_z$ for all models and tasks.
	
	\vspace{.1cm}
	\noindent \textbf{Adversarial Graph Autoencoder Model. \ }
	The equation for training the encoder model with Discriminator $\mathcal{D(\mathbf{Z})}$ can be written as follows: 
	\begin{equation}
	\label{eq:objective}
	\min_{\mathcal{G}} \max_{\mathcal{D}} \mathbbm{E}_{\mathbf{z}\sim p_{z} }[\text{log} \mathcal{D(\mathbf{Z})}] + \mathbbm{E}_{\mathbf{x}\sim p(\mathbf{x})}[\text{log}(1 - \mathcal{D}(\mathcal{G}(\mathbf{X, A})))]
	\end{equation}
	where $\mathcal{G}(\mathbf{X, A})$ and $\mathcal{D(\mathbf{Z})}$ indicate the generator and discriminator explained above.

	\begin {algorithm}[tpb]
	\begin{small}
		\caption {\small Adversarially Regularized Graph Embedding}
		\label{alg:ARGE}
		\begin {algorithmic}[1]
		\REQUIRE
		\leavevmode \\
		$\mathbf{G}=\{\mathbf{V}, \mathbf{E}, \mathbf{X}\}$: a Graph with links and features;\\
		$T$: the number of iterations;\\
		$K$: the number of steps for iterating discriminator;\\
		$d$: the dimension of the latent variable
		\ENSURE
		$\mathbf{Z} \in \mathbbm{R}^{n \times d}$ \\
		\FOR{iterator = 1,2,3, $\cdots\cdots$, $T$}
		\STATE Generate latent variables matrix $\mathbf{Z}$ through Eq.(\ref{eq:latent});
		\FOR{k = 1,2, $\cdots\cdots$, $K$}
		\STATE Sample $m$ entities \{$\mathbf{z}^{(1)}$, \dots, $\mathbf{z}^{(m)}$\} from latent  matrix $\mathbf{Z}$
		\STATE Sample $m$ entities \{$\mathbf a^{(1)}$, \dots, $\mathbf a^{(m)}$\} from the prior distribution $p_{z}$
		\STATE Update the discriminator with its stochastic gradient:
		\begin{equation*}
		\bigtriangledown \frac{1}{m}\sum^m_{i=1}[\text{log~}\mathcal{D}(\mathbf a^{i}) + \text{log~}(1 - \mathcal{D}( \mathbf{z}^{(i)}))]
		\end{equation*}
		\ENDFOR
		\STATE Update the graph autoencoder with its stochastic gradient by  Eq. (\ref{eq:autoencoder_obj0}) for ARGA or Eq. (\ref{eq:autoencoder_obj}) for ARVGA;
		\ENDFOR
		\RETURN $\mathbf{Z} \in \mathbbm{R}^{n \times d}$ 
		\end {algorithmic}
	\end{small}
	\end {algorithm}
	
	
	\begin{figure*}
		\centering
		\includegraphics[ width=.85\linewidth]{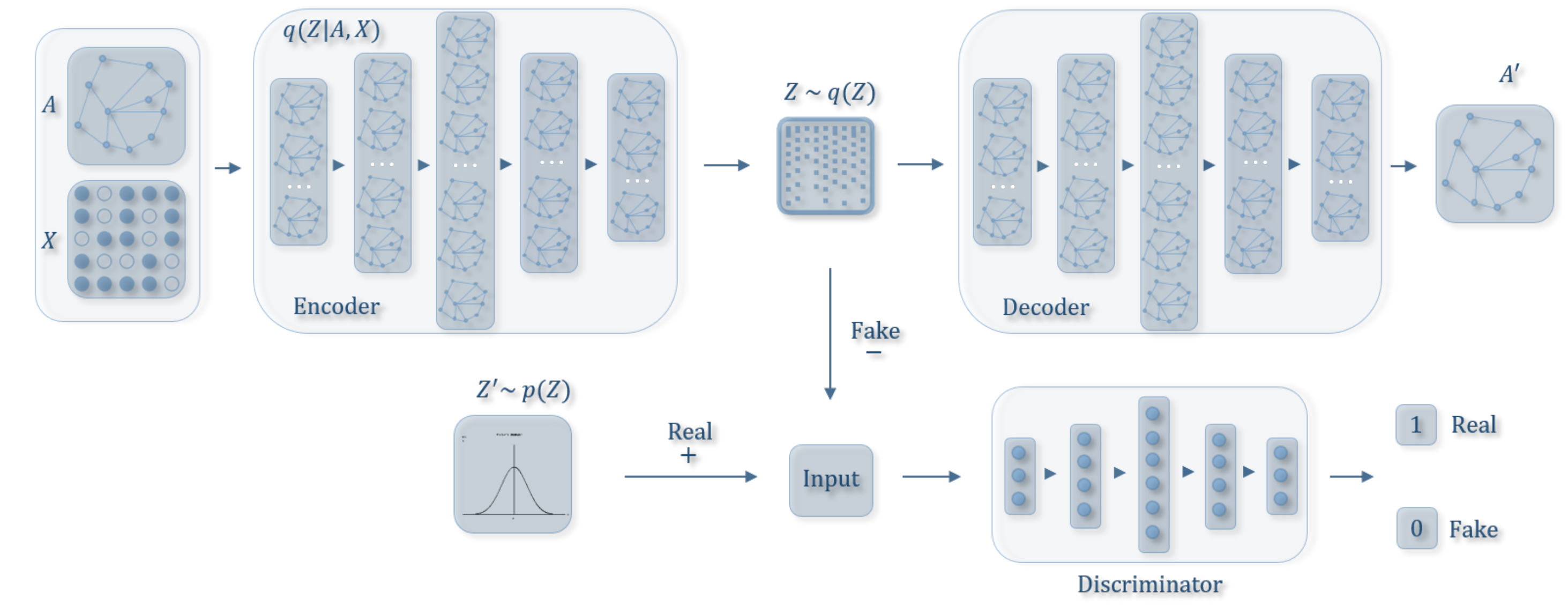}
		\caption{The architecture of adversarially regularized graph autoencoder with a graph convolutional decoder (\textbf{ARGA\_GD}) to reconstruct the topological structure $\mathbf{A}$.The upper tier is a standard graph convolutional autoencoder. The decoder employs the graph convolutional networks. The lower tier keeps the
			same with both Gaussian distribution and Uniform distribution. \textbf{ARVGA\_GD} is similar to \textbf{ARGA\_GD} except that it employs a \textit{variational} graph autoencoder in the upper tier.
		} \label{fig:introduction2}
	\end{figure*}
	
	\subsection{Algorithm Explanation}
	Algorithm \ref{alg:ARGE} is our proposed framework. Given a graph $\mathbf{G}$, step 2 gets the latent variables matrix $\mathbf{Z}$ from the graph convolutional encoder. Then we take the same number of samples from the generated $\mathbf{Z}$ and the real data distribution $p_{z}$ in step 4 and 5 respectively, to update the discriminator with the cross-entropy cost computed in step 6. After $K$ runs of training the discriminator, the graph encoder will try to confuse the trained discriminator and update itself with the generated gradient in step 8. 
	We can update Eq. (\ref{eq:autoencoder_obj0}) to train the \textbf{adversarially regularized graph autoencoder (ARGA),} or Eq. (\ref{eq:autoencoder_obj}) to train the \textbf{adversarially regularized variational graph autoencoder (ARVGA)}, respectively. Finally, we will return the graph embedding $\mathbf{Z} \in \mathbbm{R}^{n \times d}$  in step 9.

	\begin{figure*}
		\centering
		\includegraphics[ width=.85\linewidth]{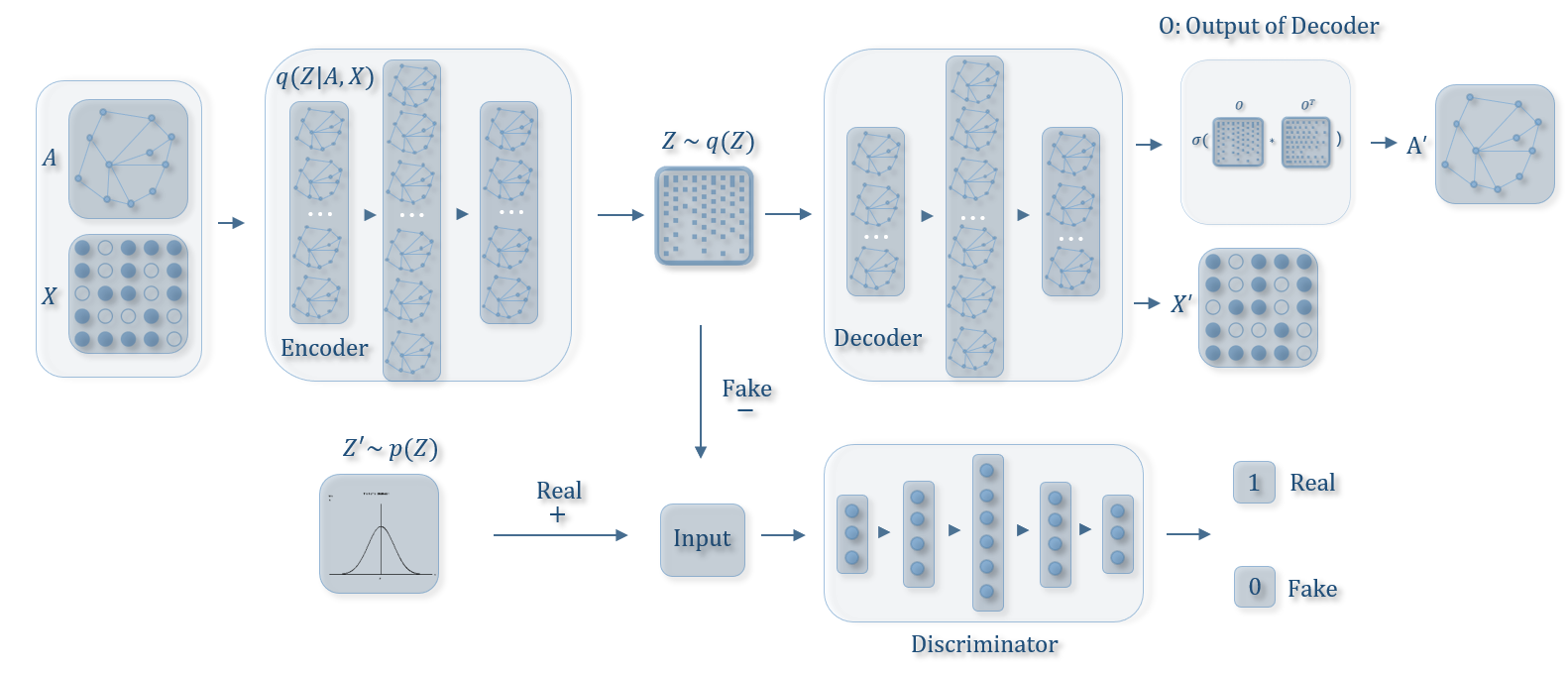}
		\caption{The architecture of the \textbf{ARGA\_AX}  which simultaneously reconstructs the graph topological structure $\mathbf{A}$ and the node content matrix $\mathbf{X}$.  The lower tier keeps the same, and we also exploit the variational version of the \textbf{ARVGA\_AX}.
		} \label{fig:introduction_new}
	\end{figure*}

	\subsection{Decoder Variations} \label{sec:variation}
	In ARGA and ARVGA models, the decoder is merely a link prediction layer which performs as a dot product of the embedding $\textbf{Z}$. In practice, the decoder can also be a graph convolutional layer or a combination of  link prediction layer and graph convolutional decoder layer.
	
	\vspace{.1cm}
	\noindent \textbf{GCN Decoder for Graph Structure Reconstruction (ARGA\_GD)}
	We have modified the encoder by adding two graph convolutional layers to reconstruct the graph structure. This variant of approach is named ARGA\_GD. Fig. \ref{fig:introduction2} demonstrates the architecture of ARGA\_GD. In this approach, the input of the decoder will be the embedding from the encoder, and the graph convolutional decoder is constructed as follows:
	\begin{eqnarray}
	\mathbf Z_{D} = f_{\text{linear}}(\mathbf{Z},\mathbf{A} | \mathbf W_{D}^{(1)}).\\
	\mathbf O = f_{\text{linear}}(\mathbf {Z}_{D},\mathbf{A} | \mathbf W_{D}^{(2)}).
	\end{eqnarray} 
    where $\mathbf{Z}$ is the embedding learned from the graph encoder while $ \mathbf Z_{D}$ and $\mathbf{O}$ are the outputs from the first and second layer of the graph decoder. The number of the horizontal dimension of $\mathbf{O}$ is equal to the number of nodes. Then we calculate the reconstruction error as follows:
	
	\begin{equation}
	\mathcal{L}_{ARGA\_GD} = \mathbbm{E}_{q(\mathbf{O|(X,A)})}[\text{log}~ p(\mathbf{A}|\mathbf{O})] 
	\end{equation}
	
	\vspace{.1cm}
	\noindent \textbf{GCN Decoder for both Graph Structure and Content Information Reconstruction (ARGA\_AX)}
	We have further modified our graph convolutional decoder to reconstruct both the graph structure $\mathbf{A}$ and content information $\mathbf{X}$. The architecture is illustrated in Fig \ref{fig:introduction_new}. We fixed the dimension of second graph convolutional layer with the same number of the features associated with every node, thus the output from the second layer $\mathbf{O}\in \mathbbm{R}^{n \times f} \ni \mathbf{X}$. In this case, the reconstruction loss is composed of two errors. First, the reconstruction error of graph structure can be minimized as follows:
	\begin{equation}
	\mathcal{L}_{A} = \mathbbm{E}_{q(\mathbf{O|(X,A)})}[\text{log}~ p(\mathbf{A}|\mathbf{O})] \label{eq:reconstruction_A},
	\end{equation}
	Then the reconstruction error of node content can be minimized with a similar formula:
	\begin{equation}
	\mathcal{L}_{X} = \mathbbm{E}_{q(\mathbf{O|(X,A)})}[\text{log}~ p(\mathbf{X}|\mathbf{O})] \label{eq:reconstruction_X}.
	\end{equation}
	The final reconstruction error is the sum of the reconstruction error of graph structure and node content:
	\begin{equation}
	\mathcal{L}_{0} = \mathcal{L}_{A} + \mathcal{L}_{X}.
	\end{equation}

	\section{Experiments}\label{sec:experiments}

	We report our results on both link prediction and node clustering tasks. The benchmark graph datasets used in the paper, Cora \cite{lu2003link}, Citeseer \cite{sen2008collective} and Pubmed \cite{namata2012query}, are summarized in table 1. 
	Each dataset consists of scientific publications as nodes and citation relationships as edges. The features are unique words  in each document. 
	
	\begin{table}[htpb]
\small
  \centering
  \setlength\tabcolsep{1.9pt}
  \caption{Real-world Graph Datasets Used in the Paper}
    \begin{tabular}{lcccccl}
    \toprule
    Data Set & \# Nodes & \# Links   & \# Content Words & \# Features\\
    \midrule
    
    Cora   &  2,708 & 5,429   & 3,880,564 & 1,433  \\
    Citeseer & 3,327 & 4,732   & 12,274,336 & 3,703  \\
    PubMed & 19,717 & 44,338  & 9,858,500   & 500 \\
    \bottomrule
    \end{tabular}%
  \label{tab:datasets}%
\end{table}%
	
	\begin{table*}
\small
  \centering
  \caption{Results for Link Prediction. $\text{GAE}^*$ and $\text{VGAE}^*$ are variants of GAE and VGAE, which only explore topological structure, i.e., $\mathbf X=\mathbf I$. }
\begin{tabular}{ccccccc}
\hline
 Approaches & \multicolumn{2}{c}{Cora} & \multicolumn{2}{c}{Citeseer}& \multicolumn{2}{c}{PubMed} \\
& AUC & AP& AUC & AP& AUC & AP\\
SC &	84.6 $\pm$  0.01 &	88.5 $\pm$ 0.00 &	80.5 $\pm$  0.01 &	85.0 $\pm$  0.01 &	84.2 $\pm$  0.02 &	87.8 $\pm$  0.01\\
DW &	83.1 $\pm$  0.01 &	85.0 $\pm$  0.00 &	80.5 $\pm$ 0.02 &	83.6 $\pm$  0.01 &	84.4 $\pm$  0.00 &	84.1 $\pm$  0.00\\
      \midrule
$\text{GAE}^*$ &	84.3 $\pm$  0.02 &	88.1 $\pm$  0.01 &	78.7 $\pm$  0.02 &	84.1 $\pm$  0.02 &	82.2 $\pm$  0.01 &	87.4 $\pm$  0.00\\
$\text{VGAE}^*$ &	84.0 $\pm$  0.02 &	87.7 $\pm$  0.01 &	78.9 $\pm$  0.03 &	84.1 $\pm$  0.02 &	82.7 $\pm$  0.01 &	87.5 $\pm$  0.01\\
GAE &	91.0 $\pm$  0.02 &	92.0 $\pm$  0.03 &	89.5 $\pm$  0.04 &	89.9 $\pm$  0.05 &	96.4 $\pm$  0.00 &	96.5 $\pm$  0.00\\
VGAE &	91.4 $\pm$  0.01 &	92.6 $\pm$  0.01 &	90.8 $\pm$  0.02 &	92.0 $\pm$  0.02 &	94.4 $\pm$  0.02 &	94.7 $\pm$  0.02\\
        \midrule
\textbf{ARGA} &	92.4 $\pm$  0.003 &	\textbf{93.2 $\pm$  0.003} &	91.9 $\pm$  0.003 &	93.0$\pm$  0.003 &	\textbf{96.8 $\pm$  0.001} &	\textbf{97.1 $\pm$  0.001}\\
\textbf{ARVGA} &	\textbf{92.4 $\pm$  0.004} &	92.6 $\pm$  0.004 &	\textbf{92.4 $\pm$  0.003} &	\textbf{93.0 $\pm$  0.003} &	96.5$\pm$  0.001 &	96.8$\pm$  0.001\\
\midrule
\textbf{$ARGA\_DG$} &	77.9 $\pm$  0.003 &	78.9 $\pm$  0.003 &	74.4 $\pm$  0.003 &	76.2$\pm$  0.003 &	95.1 $\pm$  0.001 &	95.2 $\pm$  0.001\\
\textbf{$ARVGA\_DG$} &	88.0 $\pm$  0.004 &	87.9 $\pm$  0.004 &	89.7 $\pm$  0.003 &	90.5 $\pm$  0.003 &	93.2$\pm$  0.001 &	93.6 $\pm$  0.001\\
\midrule
\textbf{$ARGA\_AX$} &	91.3 $\pm$  0.003 &	91.3 $\pm$  0.003 &	91.9 $\pm$  0.003 &	93.4$\pm$  0.003 &	96.6 $\pm$  0.001 &	96.7 $\pm$  0.001\\
\textbf{$ARVGA\_AX$} &	90.2 $\pm$  0.004 &	89.2 $\pm$  0.004 &	89.8 $\pm$  0.003 &	90.4 $\pm$  0.003 &	96.7$\pm$  0.001 &	97.1 $\pm$  0.001\\
\hline
\end{tabular}

  \label{tab:linkprediction}%
\end{table*}%

	\subsection{Link Prediction} \label{sec:link}
	\noindent\textbf{Baselines. \ } 
	\textbf{Twelve} algorithms in total are compared for the link prediction task:
	\begin{itemize}
		\item \textbf{DeepWalk} \cite{perozzi2014deepwalk} is a network representation approach which encodes social relations into a continuous vector space. 
		\item \textbf{Spectral Clustering} \cite{tang2011leveraging} is an effective approach to learn social embedding.
		\item \textbf{GAE} \cite{kipf2016variational} is the most recent autoencoder-based unsupervised framework for graph data, which naturally leverages both topological structure $\mathbf A$ and content information $\mathbf X$. \textbf{GAE}$^*$ is the version of GAE which only considers the topological information $\mathbf A$, i.e., $\mathbf X=\mathbf I$.
		\item \textbf{VGAE} \cite{kipf2016variational} is the variational graph autoencoder for graph embedding with both topological and content information. Likewise, \textbf{VGAE}$^*$ is a simplified version of VGAE which only leverages the topological information. 
		
		\item \textbf{ARGA} is our proposed adversarially regularized autoencoder algorithm which uses graph autoencoder to learn the embedding.
		
		\item \textbf{ARVGA} is  our proposed algorithm, which uses a \textit{variational} graph autoencoder to learn the embedding.
		\item \textbf{ARGA\_DG} is a variant of our proposed ARGA which takes graph convolutional layers as its decoder to reconstruct graph structure. \textbf{ARVGA\_DG} is the variational version of \textbf{ARGA\_DG}.
		\item \textbf{ARGA\_AX} is a variant of our proposed ARGA which takes graph convolutional layers as its decoder to simultaneously reconstruct graph structure and node content. \textbf{ARVGA\_AX} is the variational version of \textbf{ARGA\_AX}.
	\end{itemize}
	
	\noindent\textbf{Metric. \ } We report the results concerning AUC score  (the area under a receiver operating characteristic curve) and  average precision (AP) \cite{kipf2016variational} score which can be computed as follow:
	\begin{center}
		$\texttt{AUC} = \frac{\sum_i^1\sum_j^1 \texttt{pred}(x_i) > \texttt{pred}(y_j)}{N*M}$
	\end{center} 
	where $\texttt{pred}(\bullet)$ is the outputs from the predictor and $N$ and $M$ are the number of positive samples $x_i \in X$ and the number of negative samples $y_j \in Y$ respectively.
	We also report the Average Precision (AP) which indicates the area under the precision-recall curve:
	\begin{center}
		$\texttt{Precision} = \frac{\texttt{true\_positive}}{\texttt{true\_positive} + \texttt{false\_positive}}$
	\end{center}
	\begin{center}
		$\texttt{AP} = \frac{\sum_k \texttt{Precision}(k)}{\#\{\texttt{positive\_sample}\}}$
	\end{center}
	where $k$ is an index for the class $k$.
	
	
	We conduct each experiment 10 times and report the mean values with the standard errors as the final scores. Each dataset is separated into a training, testing set, and a validation set. The validation set contains 5\% citation edges for hyperparameter optimization, the test set holds 10\% citation edges to verify the performance, and the rest are used for training.

	\vspace{.1cm}
	\noindent\textbf{Parameter Settings. }
    For the Cora and Citeseer data sets, we train all autoencoder-related models for 200 iterations and optimize them with the Adam algorithm. Both the learning rate and discriminator learning rate are set as 0.001. As the PubMed dataset is relatively large (around 20,000 nodes), we iterate 2,000 times for adequate training with a 0.008 discriminator learning rate and 0.001 learning rate. We construct encoders with a 32-neuron hidden layer and a 16-neuron embedding layer for all the experiments and all the discriminators are built with two hidden layers(16-neuron, 64-neuron respectively). 
	For the rest of the baselines, we retain the settings described in the corresponding papers.
	

	
	

	\begin{table*}
\scriptsize
  \centering
  \setlength\tabcolsep{1.8pt}
  \setlength\extrarowheight{3.5pt}
  \caption{Algorithm Comparison}
    \begin{tabular}{cccccccccccccccccccc}
    \toprule
    \textbf{ } & K-means & Spectral & BigClam & GraphEncoder & DeepWalk & DNGR & Circles & RTM & RMSC & TADW & $\text{GAE}^*$ & $\text{VGAE}^*$ & GAE & \textbf{ARGA} & \textbf{$ARGA\_DG$} &  \textbf{$ARGA\_AX$} \\
    \midrule
Content     & $\bigstar$ &   &   &   &   &   & $\bigstar$ & $\bigstar$ & $\bigstar$ & $\bigstar$ &   &   & $\bigstar$ & $\bigstar$ & $\bigstar$ & $\bigstar$  \\
\hdashline
Structure   &   & $\bigstar$ & $\bigstar$ & $\bigstar$ & $\bigstar$ & $\bigstar$ & $\bigstar$ & $\bigstar$ & $\bigstar$ & $\bigstar$ & $\bigstar$ & $\bigstar$ & $\bigstar$ & $\bigstar$ & $\bigstar$ & $\bigstar$  \\
\hdashline
Adversarial &   &   &   &   &   &   &   &   &   &   &   &   &   & $\bigstar$ & $\bigstar$ & $\bigstar$  \\
\hdashline
GCN encoder &   &   &   &   &   &   &   &   &   &   &   &   &  $\bigstar$ & $\bigstar$ & $\bigstar$ & $\bigstar$  \\
\hdashline
GCN dncoder &   &   &   &   &   &   &   &   &   &   &   &   &   &   & $\bigstar$ & $\bigstar$  \\
\hdashline
Recover A   &   &   &   & $\bigstar$ & $\bigstar$ & $\bigstar$ &   &   &   &   & $\bigstar$ & $\bigstar$ & $\bigstar$ & $\bigstar$ & $\bigstar$ & $\bigstar$  \\
\hdashline
Recover X   &   &   &   &   &   &   &   &   &   &   &   &   &   &   &   & $\bigstar$  \\

        \bottomrule
    \end{tabular}%
  \label{tab:algorithm_comparison}%
\end{table*}%
	
	\vspace{.1cm}
	\noindent\textbf{Experimental Results. \ }
	The details of the experimental results on the link prediction are shown in Table 2. The results show that by incorporating an effective adversarial training module into our graph convolutional autoencoder, ARGA and ARVGA achieve outstanding performance: all AP and AUC scores are as higher as 92\% on all three data sets. Compared with all the baselines, ARGA increased the AP score from around 2.5\% compared with VGAE incorporating with node features,
	11\% compared with VGAE without node features; 15.5\% and 10.6\% compared with DeepWalk and Spectral Clustering respectively on the large PubMed data set.
	
	The approaches which use both node content and topological information are always straightforward to get better performance compared to those only consider graph structure. The gap between ARGA and GAE models demonstrates that regularization on the latent codes has its advantage to learn a robust embedding. The impact of various distributions, architectures of the decoder as well as the reconstructions will be discussed in Section \ref{sec:comparison}: ARGA Architectures Comparison.
	
	\begin{figure}[htpb]
		\centering
		\includegraphics[ width=0.49\linewidth]{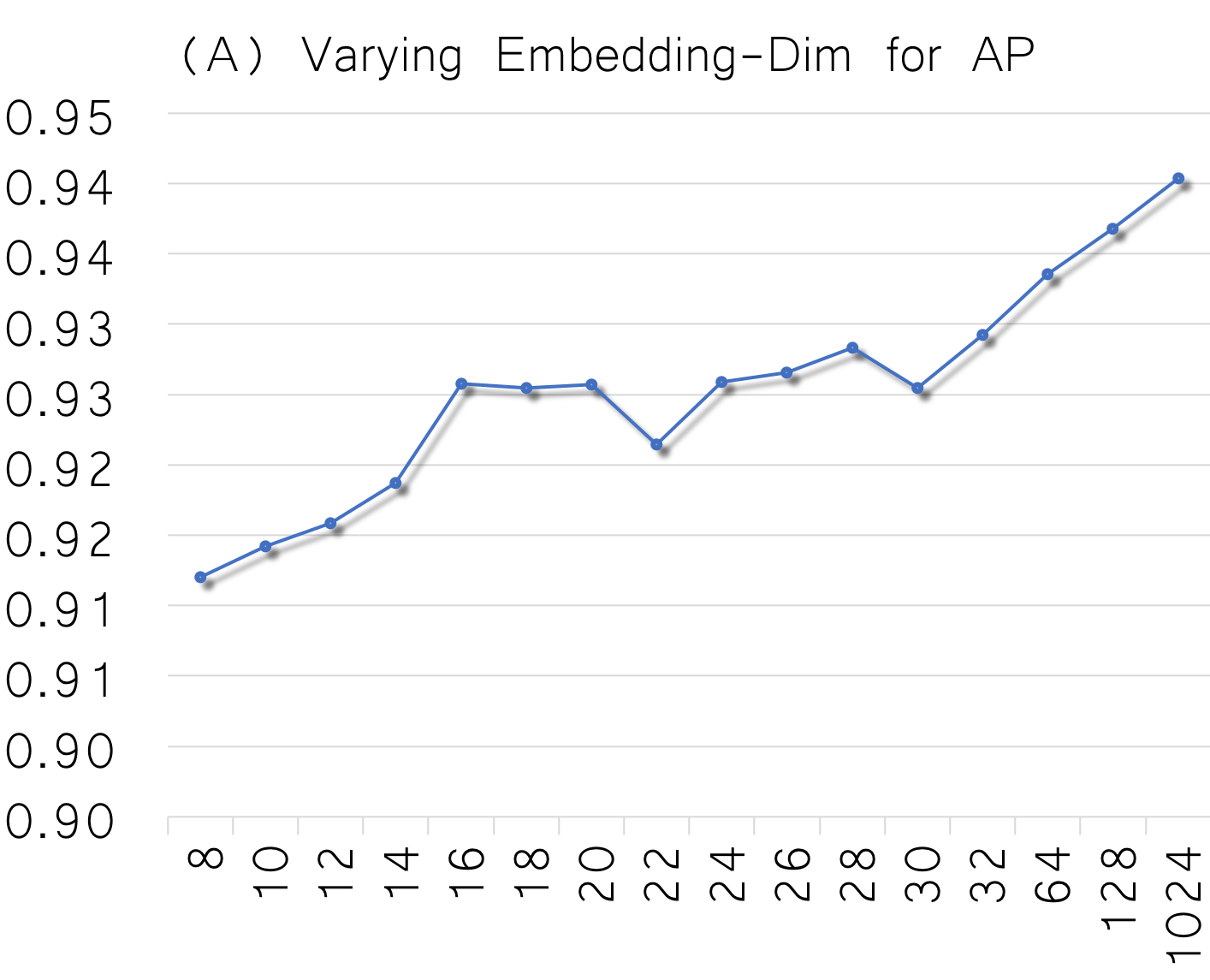}
		\includegraphics[ width=0.49\linewidth]{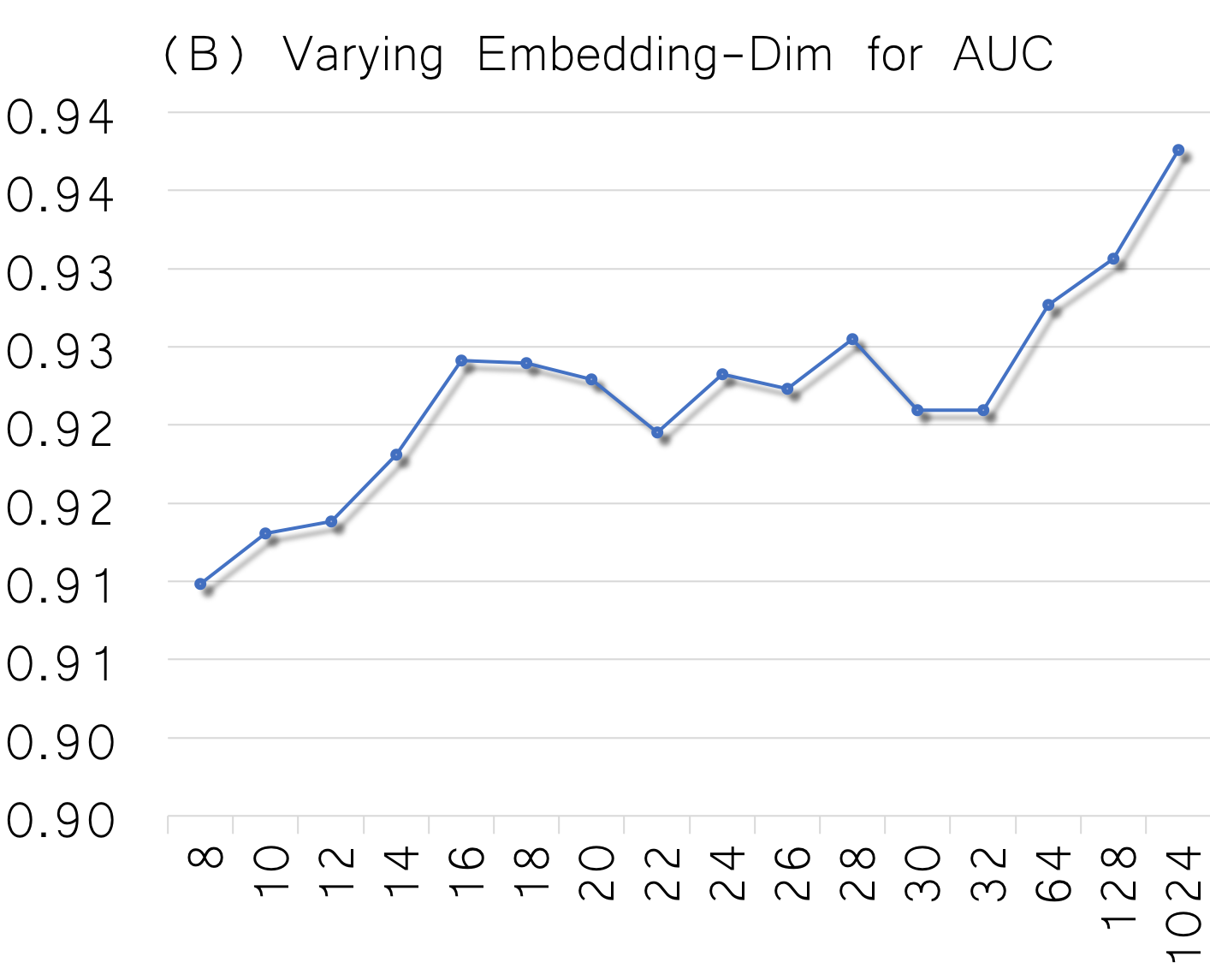}
		\caption{Average performance on different dimensions of the embedding. (A) Average Precision score; (B) AUC score.} \label{fig:embedding}
	\end{figure}
	
	\vspace{.1cm}
	\noindent\textbf{Parameter Study. \ }
	We conducted experiments on Cora dataset by varying the dimension of embedding from 8 neurons to 1024 and report the results  in Fig \ref{fig:embedding}. 
	
	The results from both Fig \ref{fig:embedding} (A) and (B) reveal similar trends: when adding the dimension of embedding from 8-neuron to 16-neuron, the performance of embedding on link prediction steadily rises; when we further increase the number of the neurons at the embedding layer to 32-neuron, the performance fluctuates, however, the results for both the AP score and the AUC score remain good.
	
	It is worth mentioning that if we continue to set more neurons, for examples, 64-neuron, 128-neuron and 1024-neuron, the performance rises dramatically.

	\subsection{Node Clustering}
	For the node clustering task, we first learn the graph embedding, and after that, we perform the K-means clustering method based on the embedding.

	\vspace{.1cm}
	\noindent\textbf{Baselines}
	We compare both embedding based approaches as well as approaches directly for graph clustering. Except for the baselines we compared for link prediction, we also include baselines which are designed for clustering.
    \textbf{Twenty} approaches in total are compared in the experiments. For a comprehensive validation, we take the algorithms which only consider one perspective of the information source, say, network structure or node content, as well as algorithms considering both factors.	

	\noindent\textbf{Node Content or Graph Structure Only:}
	\begin{enumerate}
		
		\item \textbf{K-means} is a classical method and also the foundation of many clustering algorithms.
		\item \textbf{Big-Clam} \cite{tang2011leveraging} is a community detection algorithm based on NMF.
		\item \textbf{Graph Encoder} \cite{tian2014learning} learns graph embedding for spectral graph clustering.
		\item \textbf{DNGR} \cite{cao2016deep}  trains a stacked denoising autoencoder for graph embedding. 
	\end{enumerate}
	\noindent \textbf{Both Content and Structure}
	\begin{enumerate}
		\setcounter{enumi}{4}
		\item \textbf{Circles} \cite{leskovec2012learning} is an overlapping graph clustering algorithm which treats each node as ego and builds the ego graph with the linkages between the ego's friends.
		\item \textbf{RTM} \cite{chang2009relational} learns the topic distributions of each document from both text and citation.
		\item \textbf{RMSC} \cite{xia2014robust} is a multi-view clustering algorithm which recovers the shared low-rank transition probability matrix from each view for clustering. In this paper, we treat node content and topological structure as two different views. 
		\item \textbf{TADW} \cite{yang2015network} applies matrix factorization for network representation learning.
	\end{enumerate}

	Table \ref{tab:algorithm_comparison} gives the detailed comparison of most of the baselines. For space saving, we did not list 
	the variational versions of our models. Recovering $\mathbf A$ and  $\mathbf X$ in the table demonstrates whether the model reconstructs the graph structure ($\mathbf A$) and node content ($\mathbf X$).
	Please note that we do not report the clustering results from Circle on PubMed dataset as the single experiment have been running more than three days without any outcome and error. We think this is because of the large size of the PubMed dataset (around 20,000 nodes). Note that the Circle algorithm works well on the other two datasets.

	\begin{table}[htpb]
\small
  \centering
  \caption{Clustering Results on Cora}
    \begin{tabular}{lccccccl}
    \toprule
    \textbf{Cora} & Acc & NMI & F1 & Precision & ARI\\
    \midrule
K-means & 0.492 & 0.321 & 0.368 & 0.369  & 0.230\\
Spectral & 0.367 & 0.127 & 0.318 & 0.193 &  0.031\\
BigClam  & 0.272 & 0.007 & 0.281 & 0.180 &0.001\\
GraphEncoder & 0.325 & 0.109 & 0.298 & 0.182& 0.006\\
DeepWalk  & 0.484 & 0.327 & 0.392 & 0.361  & 0.243\\
DNGR & 0.419 & 0.318 & 0.340 & 0.266  & 0.142\\
    \midrule
Circles & 0.607 & 0.404 & 0.469 & 0.501  & 0.362\\
RTM & 0.440 & 0.230 & 0.307 & 0.332 &  0.169\\
RMSC & 0.407 & 0.255 & 0.331 & 0.227 &  0.090\\
TADW & 0.560 & 0.441 & 0.481 & 0.396 & 0.332\\
    \midrule
$\text{GAE}^*$  & 0.439 & 0.291 & 0.417 & 0.453 & 0.209\\
$\text{VGAE}^*$  & 0.443 & 0.239 & 0.425 & 0.430 & 0.175\\
GAE & 0.596 & 0.429 & 0.595 & 0.596 & 0.347\\
VGAE & 0.609 & 0.436 & 0.609 & 0.609 & 0.346\\
    \midrule
\textbf{ARGA} & 0.640 & 0.449 & 0.619 & 0.646 & 0.352\\
\textbf{ARVGA} & 0.638 & 0.450 & 0.627 & 0.624 & 0.374\\
\midrule
\textbf{$ARGA\_DG$} & 0.604 & 0.425 & 0.594 & 0.600 & 0.373\\
\textbf{$ARVGA\_DG$} & 0.463 & 0.387 & 0.455 & 0.524 & 0.265\\
\midrule
\textbf{$ARGA\_AX$} & 0.597 & 0.455 & 0.579 & 0.593 & 0.366\\
\textbf{$ARVGA\_AX$} & \textbf{0.711} & \textbf{0.526} & \textbf{0.693} & \textbf{0.710} & \textbf{0.495}\\

        \bottomrule
    \end{tabular}%
  \label{tab:clustering_cora}%
\end{table}%
	\begin{table}[htbp]
\small
  \centering
  \caption{Clustering Results on Citeseer}
    \begin{tabular}{lccccccl}
    \toprule
\textbf{Citeseer} & Acc & NMI & F1 & Precision & ARI\\
\midrule K-means & 0.540 & 0.305 & 0.409 & 0.405 &0.279\\
Spectral & 0.239 & 0.056 & 0.299 & 0.179 &  0.010\\
BigClam & 0.250 & 0.036 & 0.288 & 0.182 &  0.007\\
GraphEncoder & 0.225 & 0.033 & 0.301 & 0.179 & 0.010\\DeepWalk & 0.337 & 0.088 & 0.270 & 0.248 & 0.092\\DNGR & 0.326 & 0.180 & 0.300 & 0.200 & 0.044\\
\midrule 
Circles & 0.572 & 0.301 & 0.424 & 0.409 & 0.293\\
RTM  & 0.451 & 0.239 & 0.342 & 0.349 &  0.203\\
RMSC & 0.295 & 0.139 & 0.320 & 0.204 &  0.049\\
TADW & 0.455 & 0.291 & 0.414 & 0.312 & 0.228\\
\midrule
$\text{GAE}^*$ & 0.281 & 0.066 & 0.277 & 0.315 & 0.038\\
$\text{VGAE}^*$ & 0.304 & 0.086 & 0.292 & 0.331 & 0.053\\
GAE & 0.408 & 0.176 & 0.372 & 0.418 & 0.124\\
VGAE & 0.344 & 0.156 & 0.308 & 0.349 & 0.093\\
\midrule
\textbf{ARGA} & 0.573 & 0.350 & 0.546 & 0.573 & 0.341\\
\textbf{ARVGA} & 0.544 & 0.261 & 0.529 & 0.549 & 0.245\\
\midrule
\textbf{$ARGA\_DG$} & 0.479 & 0.231 & 0.446 & 0.456 & 0.203\\
\textbf{$ARVGA\_DG$} & 0.448 & 0.256 & 0.410 & 0.496 & 0.149\\
\midrule
\textbf{$ARGA\_AX$} & 0.547 & 0.263 & 0.527 & 0.549 & 0.243\\
\textbf{$ARVGA\_AX$} & \textbf{0.581} & \textbf{0.338} & \textbf{0.525} & \textbf{0.537} & \textbf{0.301}\\
        \bottomrule
    \end{tabular}%
  \label{tab:cluatering_citeseer}%
\end{table}%
	\begin{table}[htpb]
\small
  \centering
  \caption{Clustering Results on Pubmed}
    \begin{tabular}{lccccccl}
    \toprule
\textbf{Pubmed} & Acc & NMI & F1 & Precision & ARI\\
\midrule K-means & 0.398 & 0.001 & 0.195 & 0.579 &0.002\\
Spectral & 0.403 & 0.042 & 0.271 & 0.498 &  0.002\\
BigClam & 0.394 & 0.006 & 0.223 & 0.361 &  0.003\\
GraphEncoder & 0.531 & 0.209 & 0.506 & 0.456 & 0.184\\
DeepWalk & 0.684 & 0.279 & 0.670 & 0.686 & 0.299\\
DNGR & 0.458 & 0.155 & 0.467 & 0.629 & 0.054\\
\midrule 
RTM  & 0.574 & 0.194 & 0.444 & 0.455 &  0.148\\
RMSC & 0.576 & 0.255 & 0.521 & 0.482 &  0.222\\
TADW & 0.354 & 0.001 & 0.335 & 0.336 & 0.001\\
\midrule
$\text{GAE}^*$ & 0.581 & 0.196 & 0.569 & 0.636 & 0.162\\
$\text{VGAE}^*$ & 0.504 & 0.162 & 0.504 & 0.631 & 0.088\\
GAE & 0.672 & 0.277 & 0.660 & 0.684 & 0.279\\
VGAE & 0.630 & 0.229 & 0.634 & 0.630 & 0.213\\
\midrule
\textbf{ARGA} & 0.668 & \textbf{0.305} & 0.656 & \textbf{0.699} & 0.295\\
\textbf{ARVGA} & \textbf{0.690} & 0.290 & \textbf{0.678} & 0.694 & \textbf{0.306}\\
\midrule
\textbf{$ARGA\_DG$} & 0.630 & 0.212 & 0.629 & 0.631 & 0.209\\
\textbf{$ARVGA\_DG$} & 0.630 & 0.226 & 0.632 & 0.629 & 0.212\\
\midrule
\textbf{$ARGA\_AX$} & 0.637  & 0.245  & 0.639 & 0.642 & 0.231\\
\textbf{$ARVGA\_AX$} & 0.640 & 0.239 & 0.644 & 0.639 & 0.226\\
        \bottomrule
    \end{tabular}%
  \label{tab:clustering_pubmed}%
\end{table}%
	
	\begin{figure*}
		\centering
		\includegraphics[ width=0.31\linewidth, height=4cm]{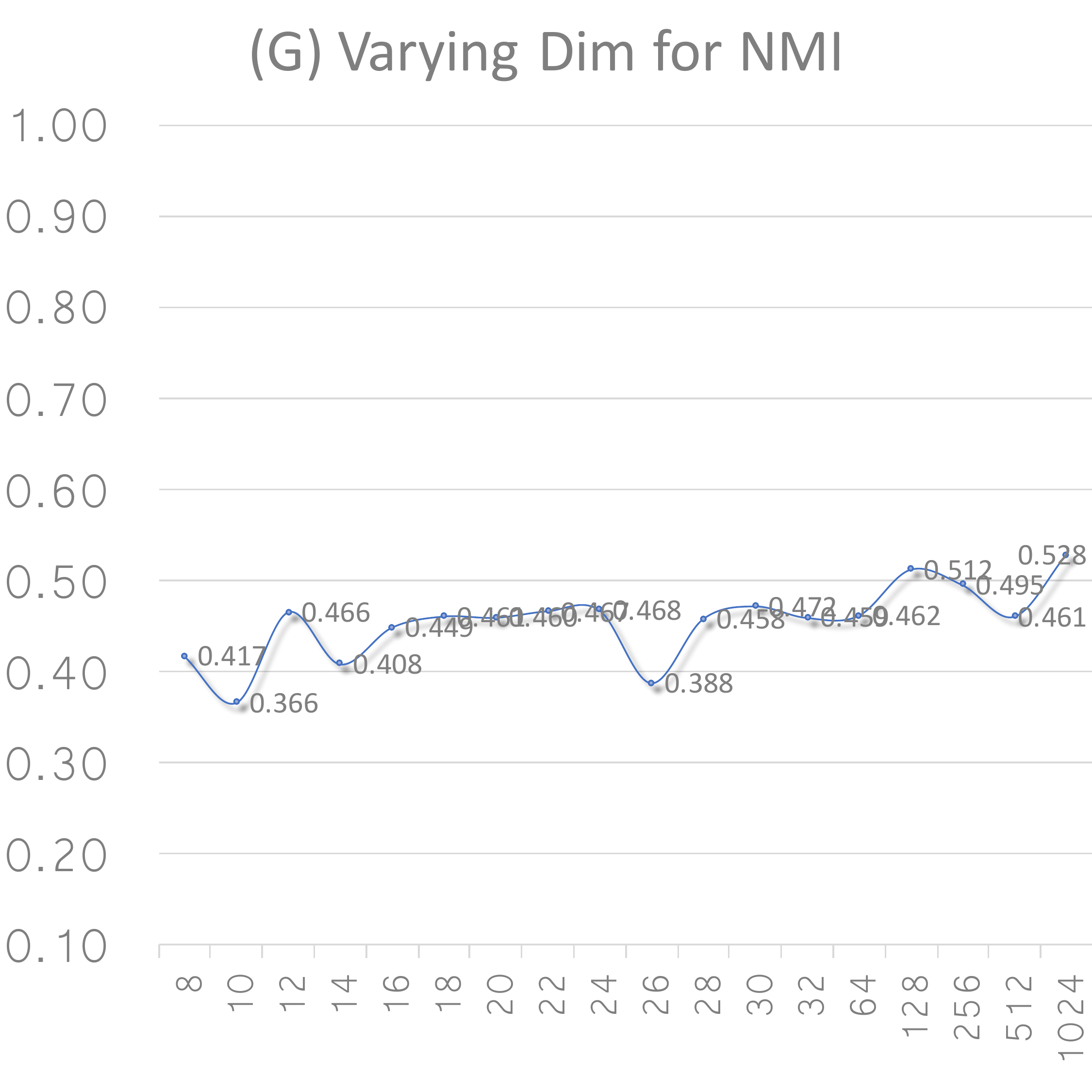}
		\includegraphics[ width=0.31\linewidth,height=4cm]{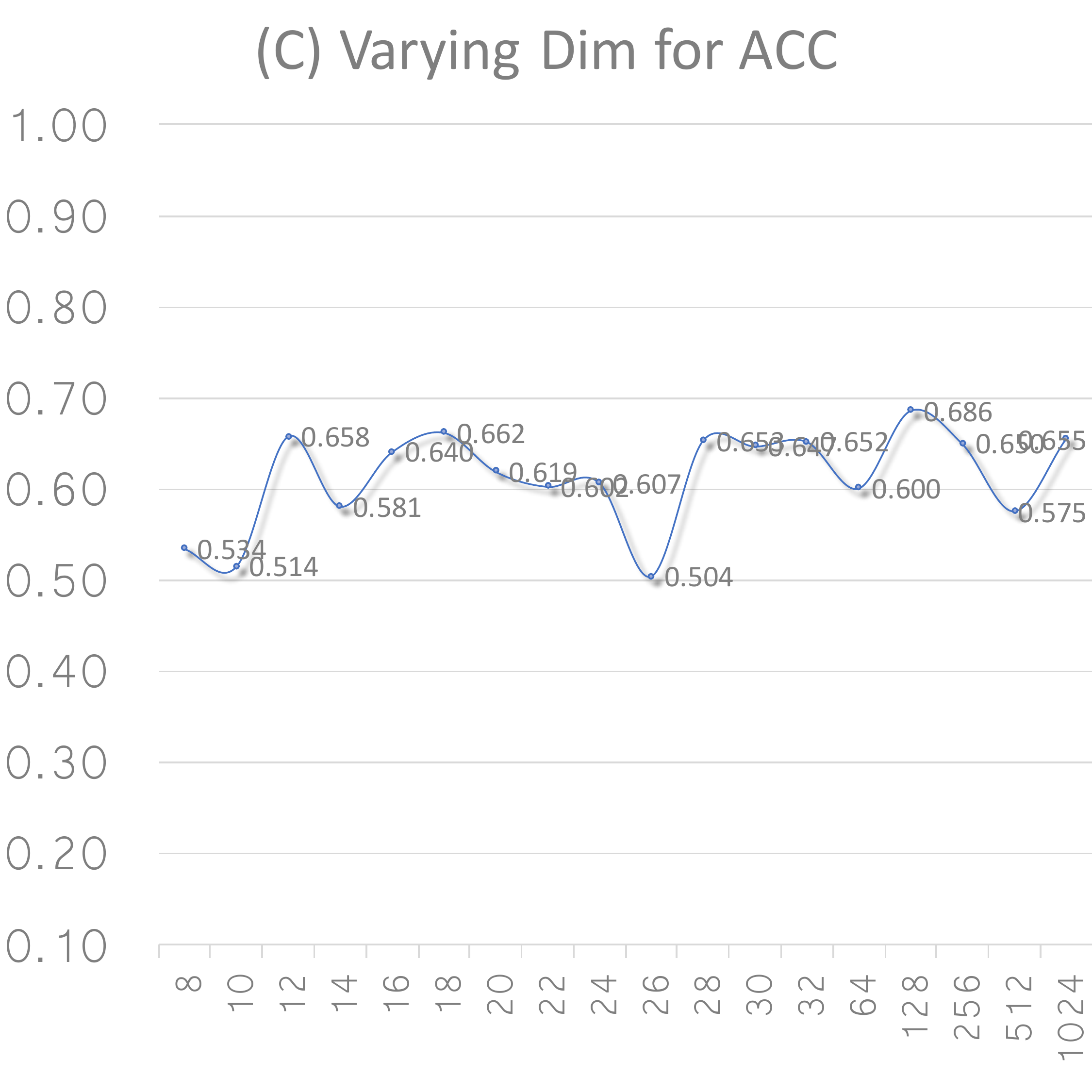}
		\includegraphics[ width=0.31\linewidth,height=4cm]{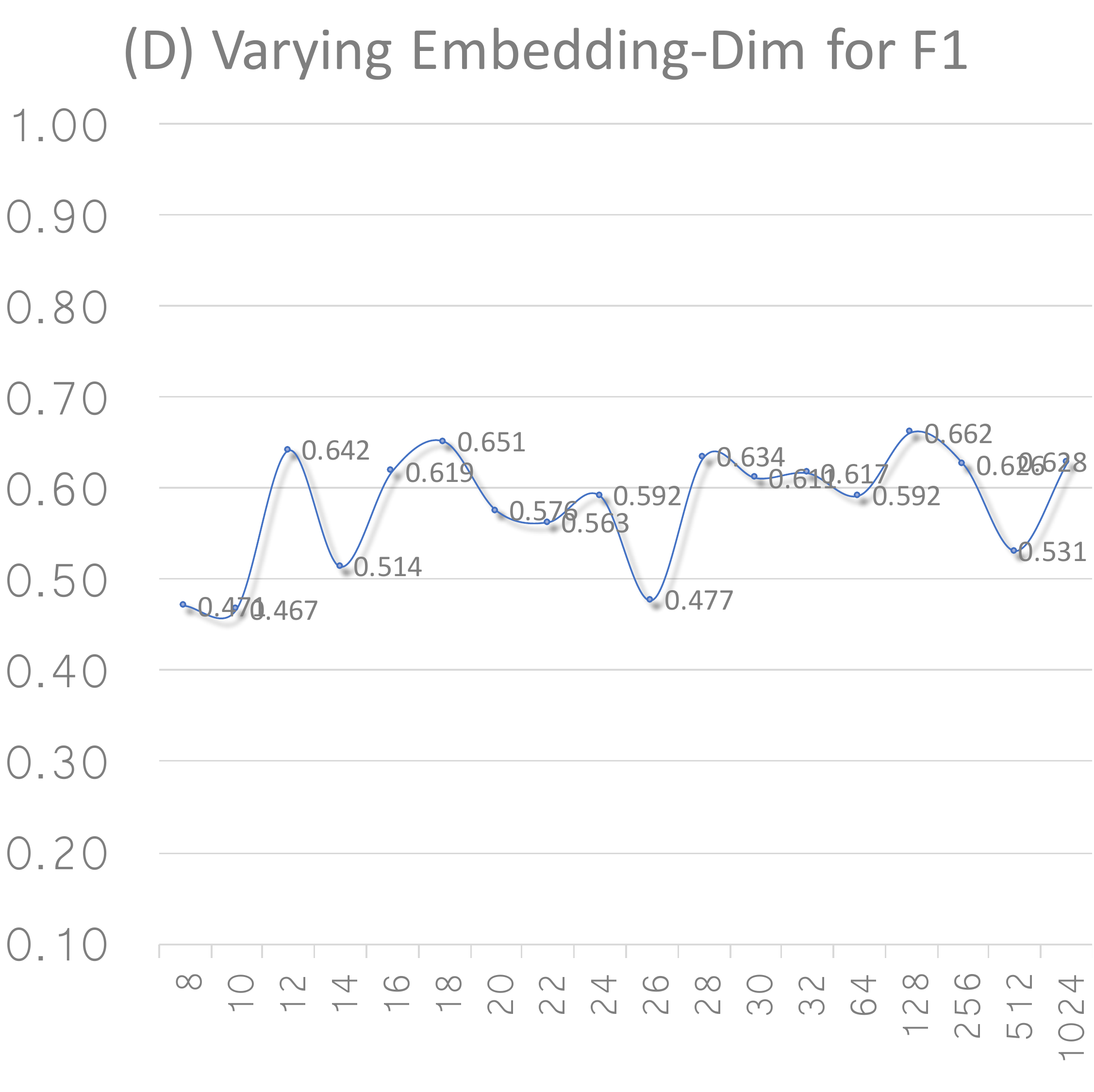}
		\includegraphics[ width=0.31\linewidth,height=4cm]{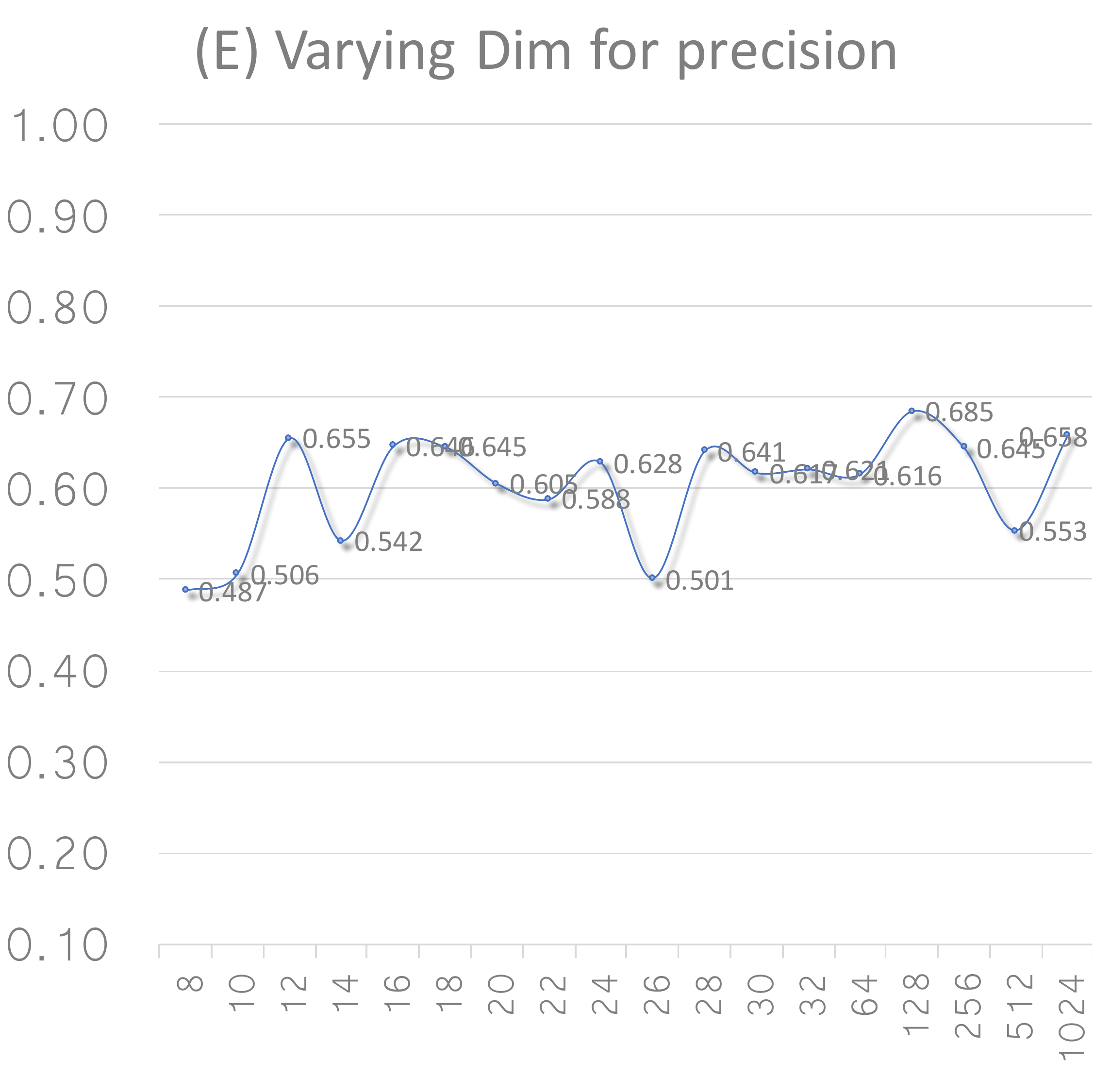}
		\includegraphics[ width=0.31\linewidth,height=4cm]{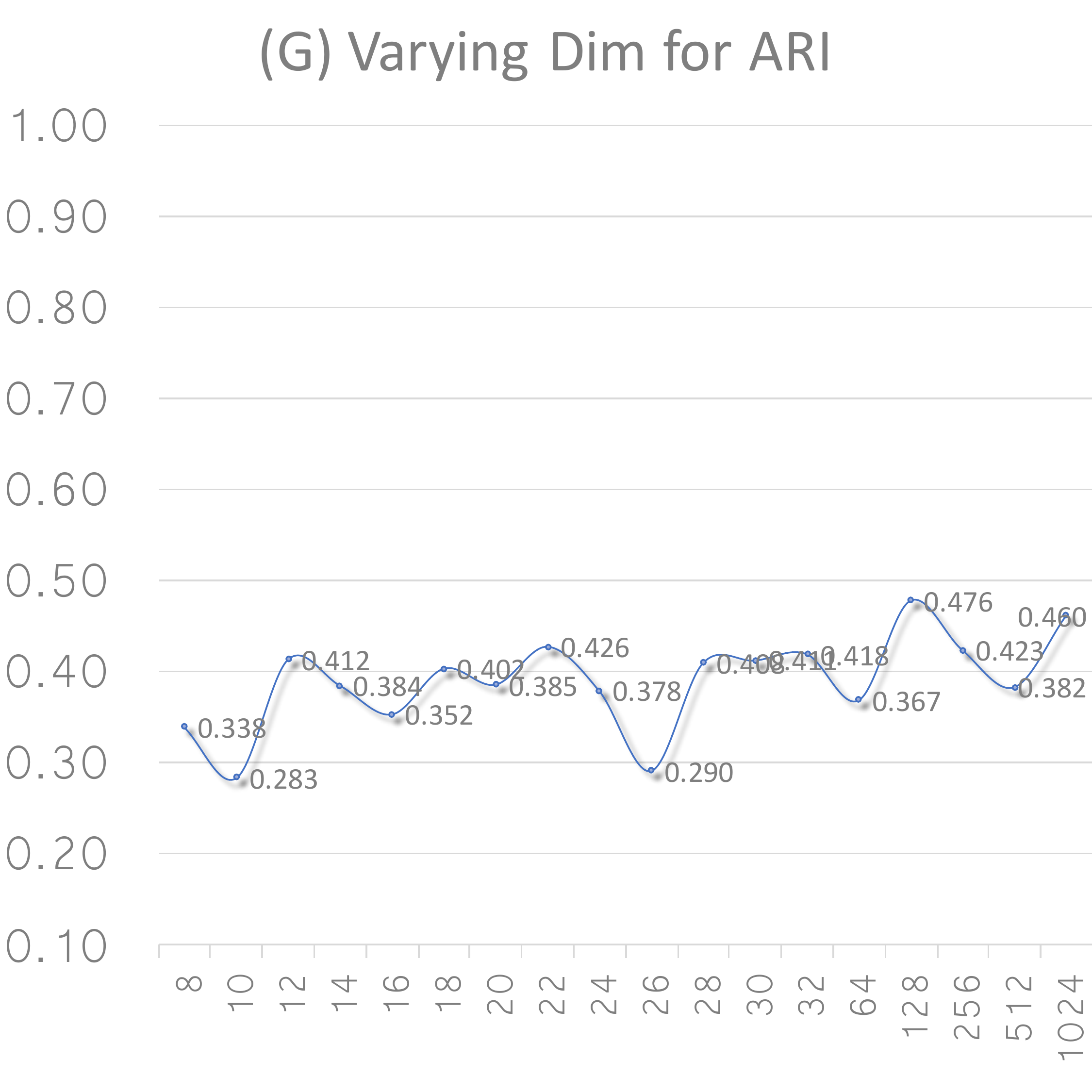}
		\includegraphics[ width=0.31\linewidth,height=4cm]{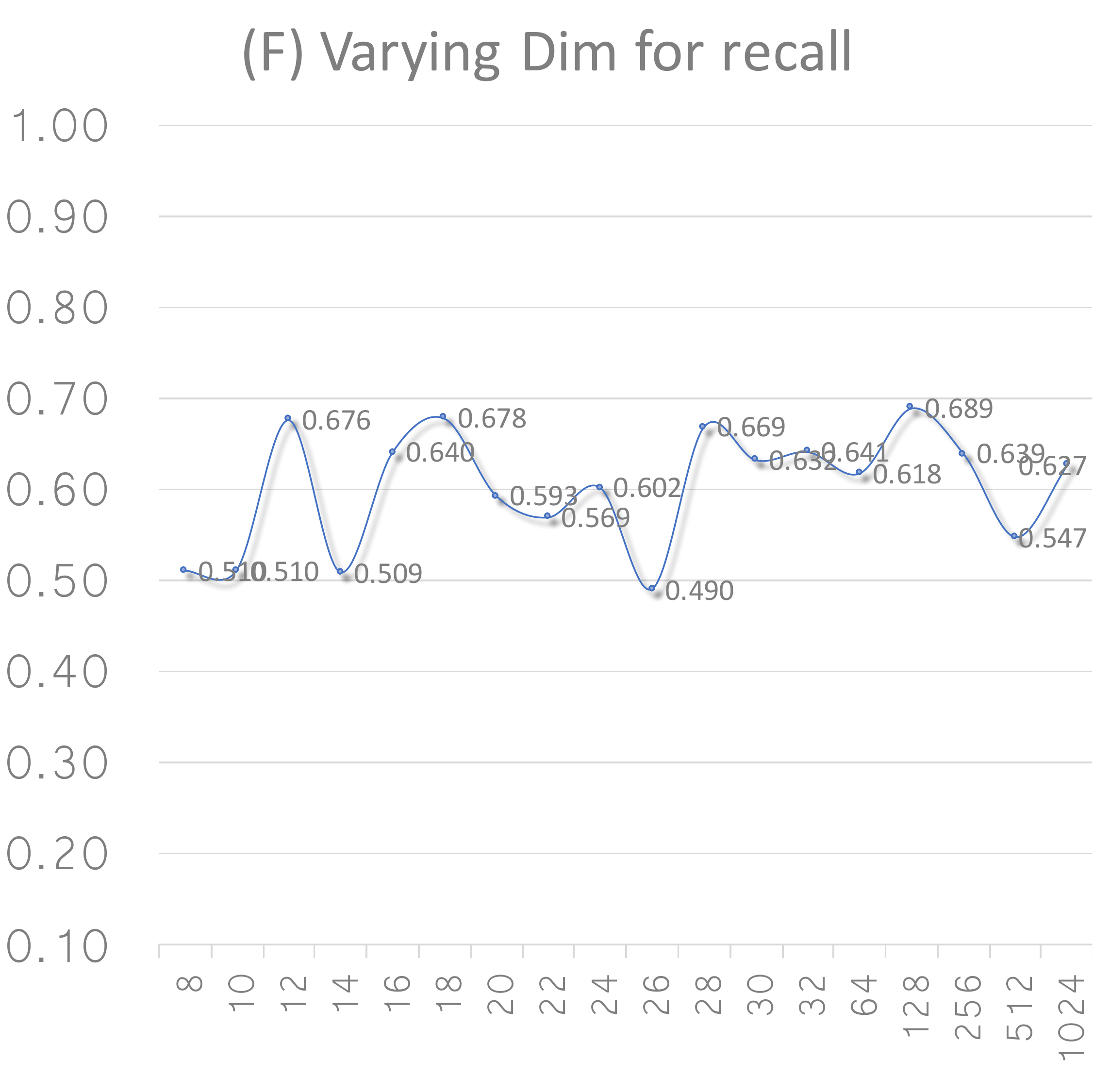}
		
		\caption{Average node clustering performance on different dimensions of the embedding. } \label{fig:embedding_clustering}
	\end{figure*}

	\vspace{.1cm}
	\noindent\textbf{Metrics: \ }
	Following \cite{xia2014robust}, we employ five metrics to validate the clustering results: accuracy (Acc), F-one score (F1), normalized mutual information (NMI), precision and average rand index (ARI).

	\vspace{.1cm}
	\noindent\textbf{Experimental Results. \ }
	The clustering results on the Cora, Citeseer and Pubmed data sets are given in table \ref{tab:clustering_cora}, table \ref{tab:cluatering_citeseer} and table \ref{tab:clustering_pubmed}. The results show that ARGA and ARVGA have achieved a dramatic improvement on all five metrics compared with all the other baselines. For instance, on Citeseer, ARGA has increased the accuracy from 6.1\% compared with K-means to 154.7\% compared with GraphEncoder; increased the F1 score from 31.9\% compared with TADW to 102.2\% compared with DeepWalk; and increased NMI from 14.8\% compared with K-means to 124.4\% compared with VGAE.
	
		\begin{figure*}
		\centering
		\includegraphics[width=0.49\linewidth, height=3.5cm]{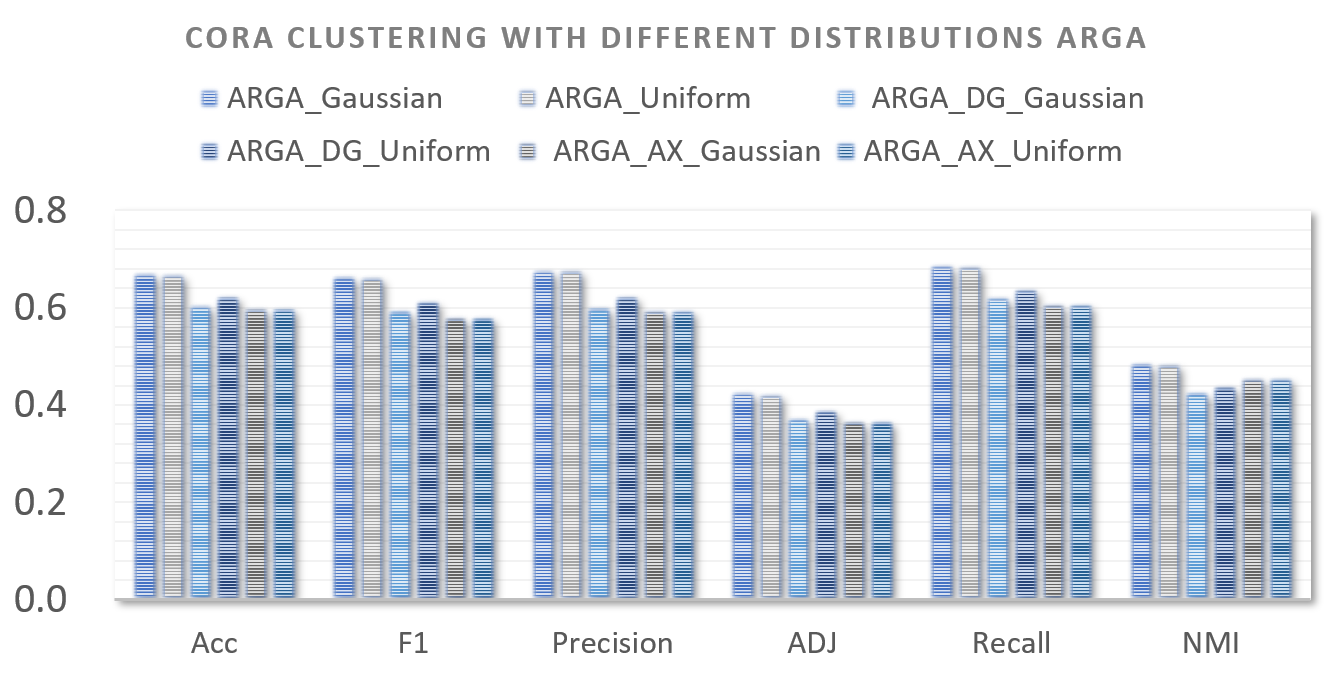}
		\includegraphics[width=0.49\linewidth, height=3.5cm]{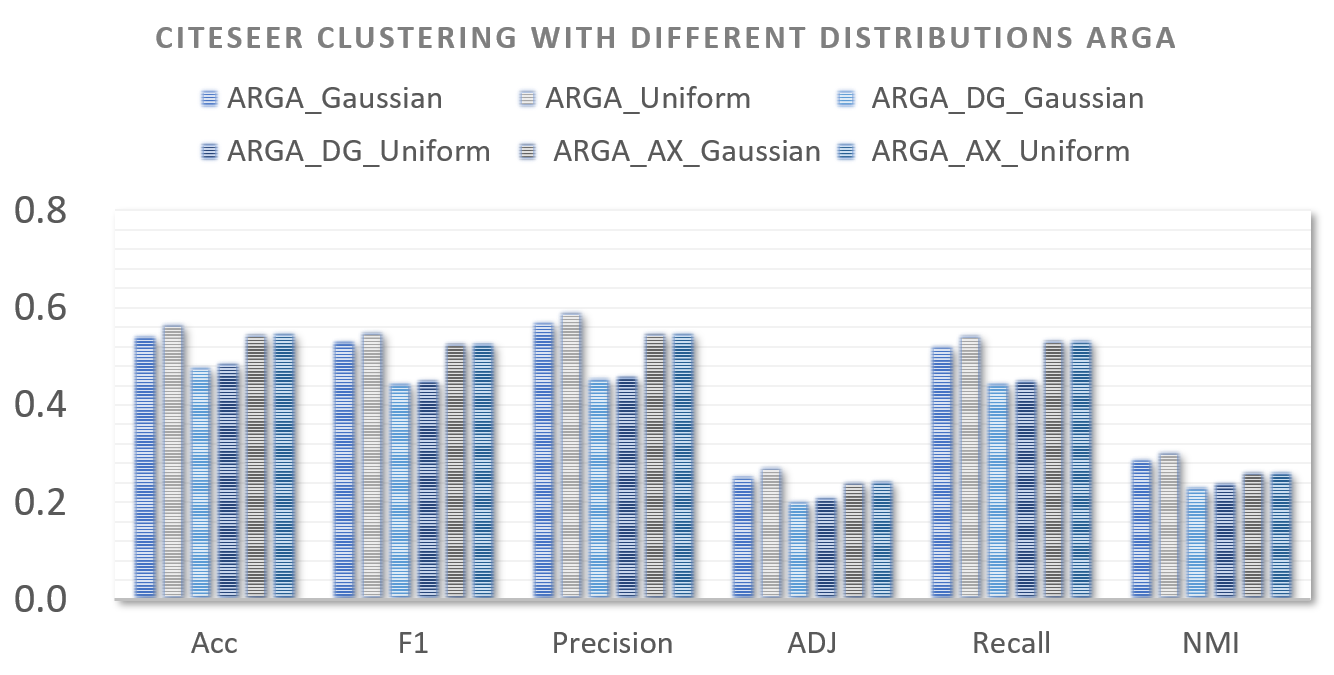}
		\includegraphics[width=0.49\linewidth, height=3.5cm]{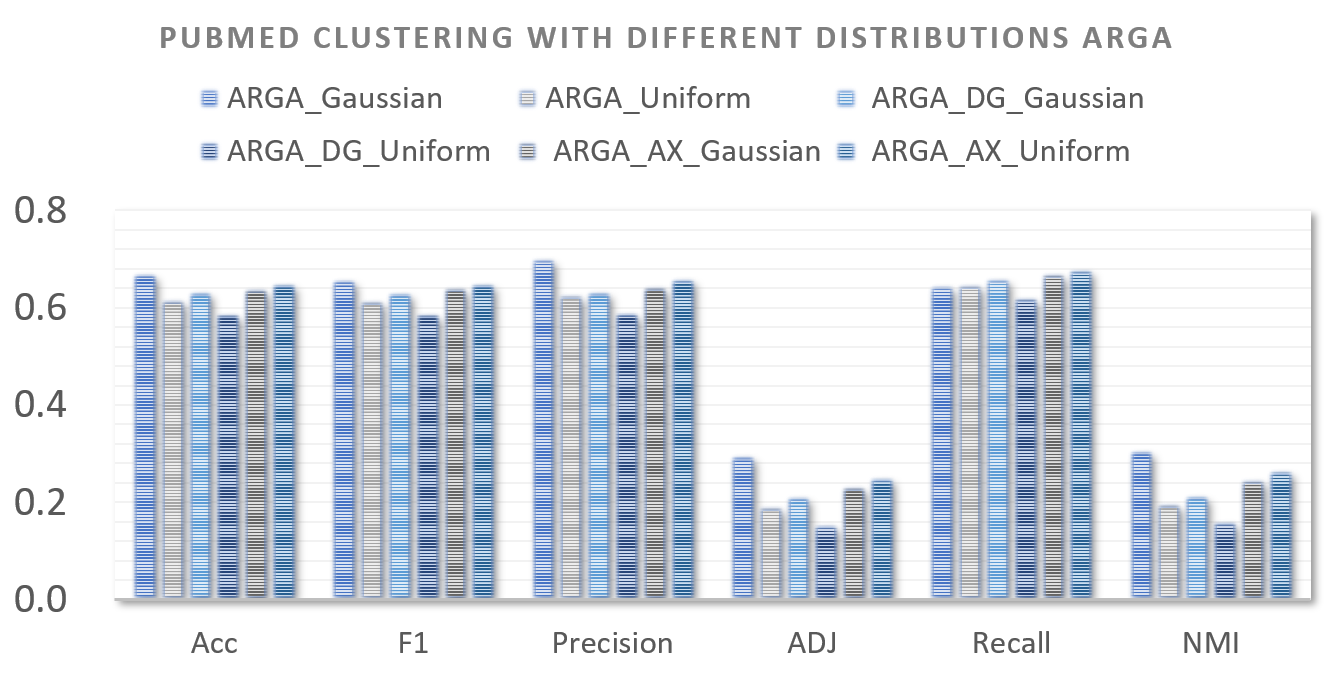}
		\includegraphics[width=0.49\linewidth, height=3.5cm]{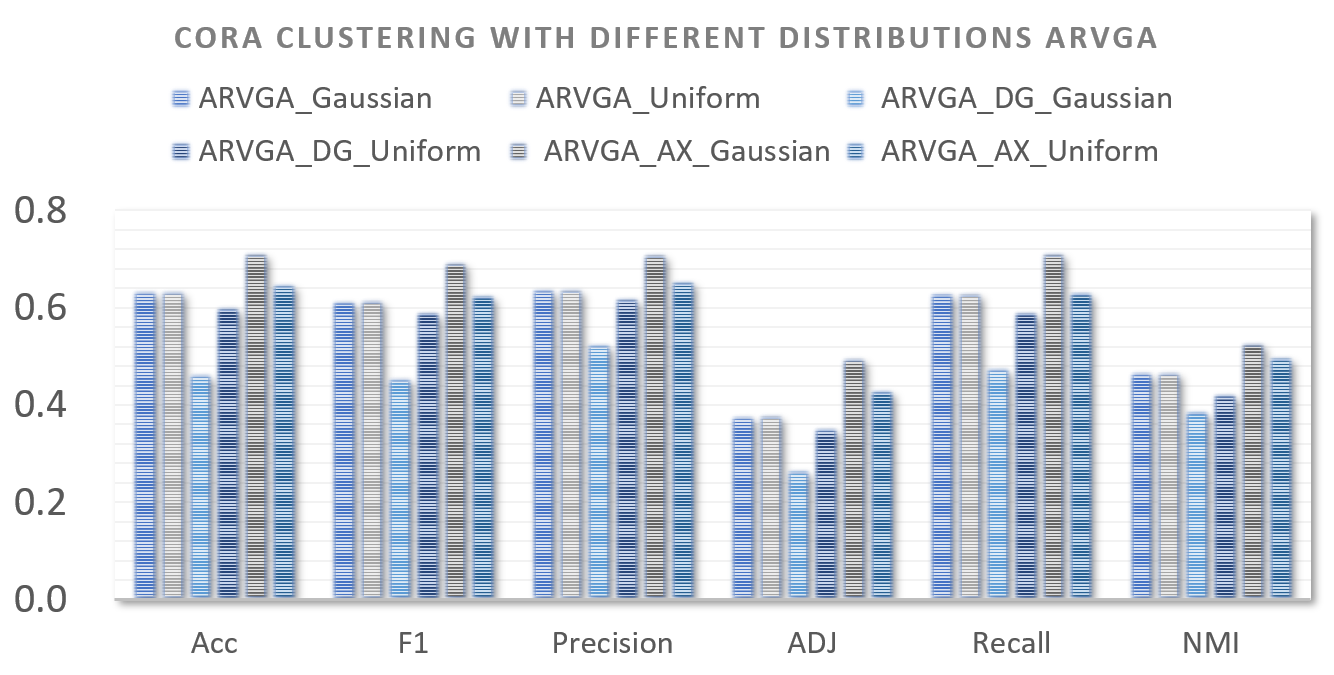}
		\includegraphics[width=0.49\linewidth, height=3.5cm]{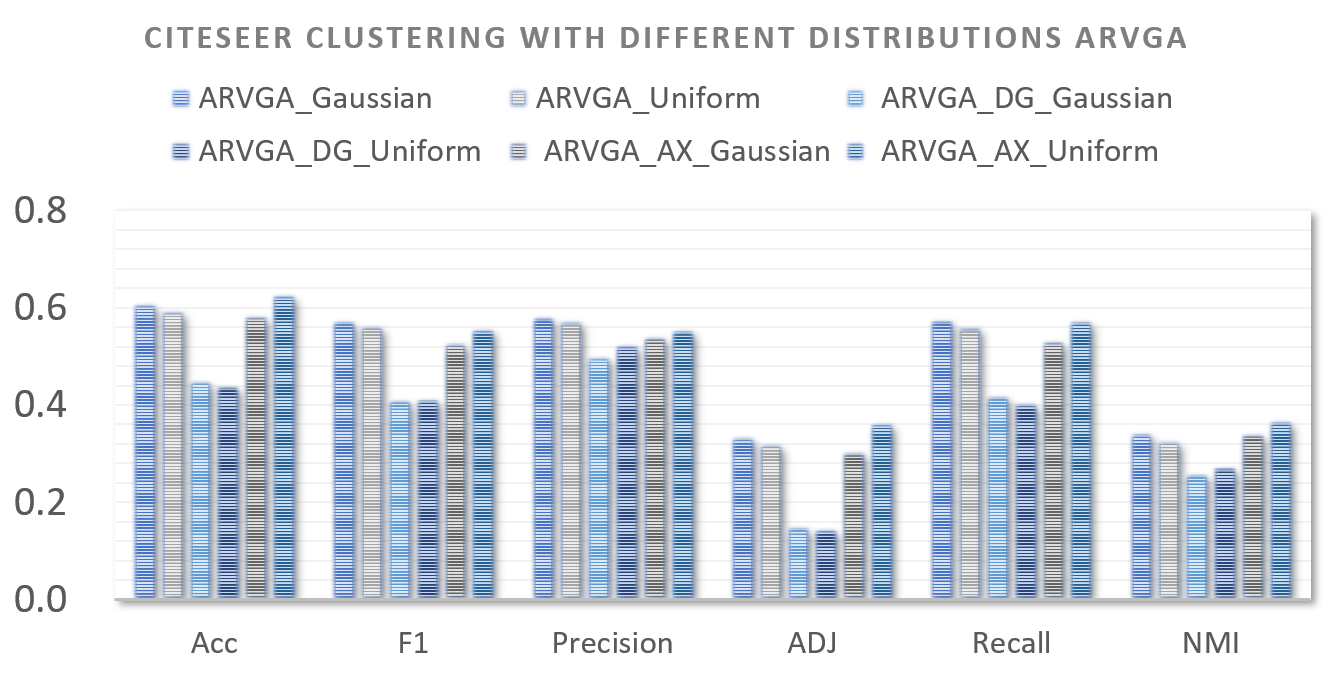}
		\includegraphics[width=0.49\linewidth, height=3.5cm]{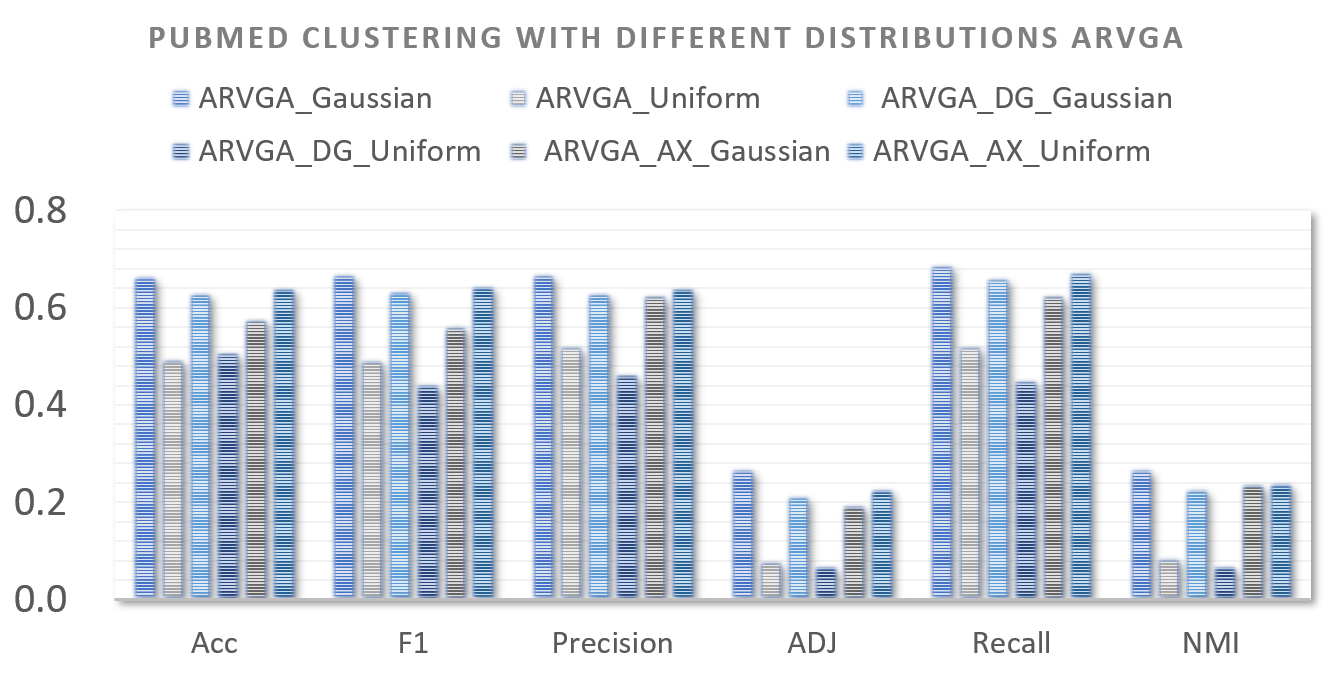}
		\caption{The ARGA related models comparison on the \textbf{clustering} task with different prior distributions.
		} \label{fig:clustering}
	\end{figure*}
	
	\begin{figure*}
		\centering
		\includegraphics[width=0.49\linewidth,height=3.5cm]{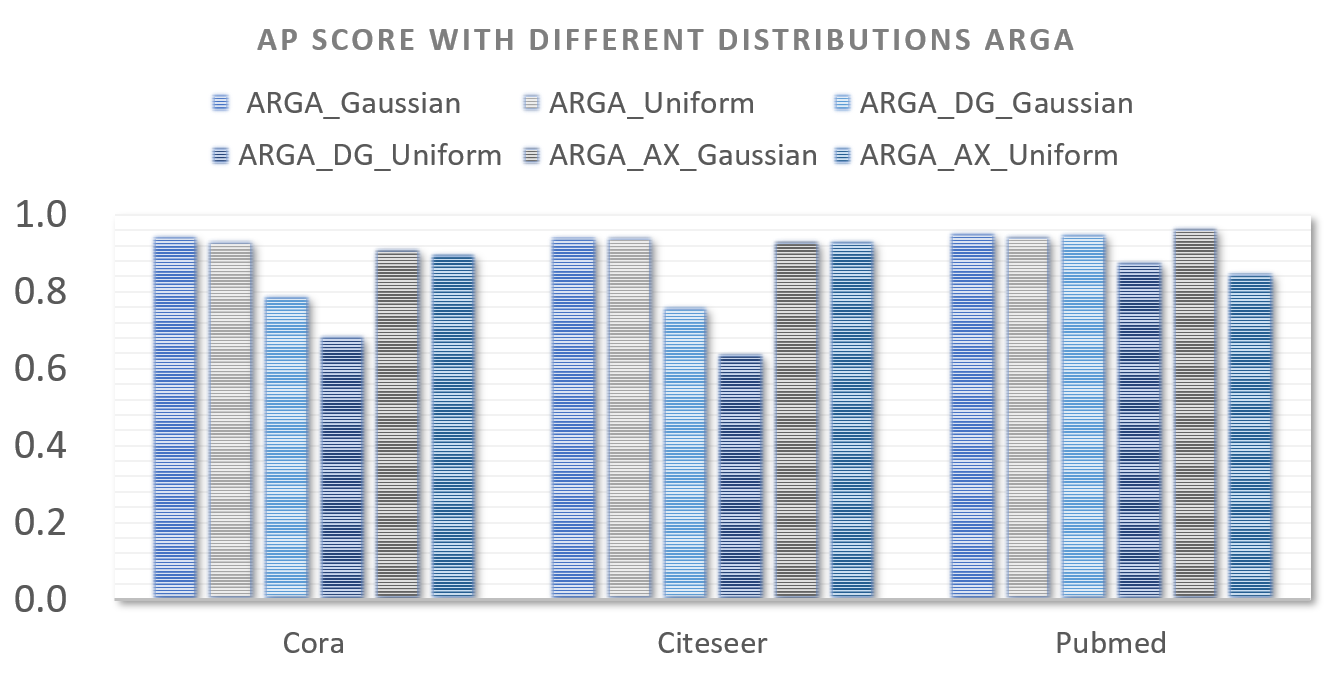}
		\includegraphics[width=0.49\linewidth,height=3.5cm]{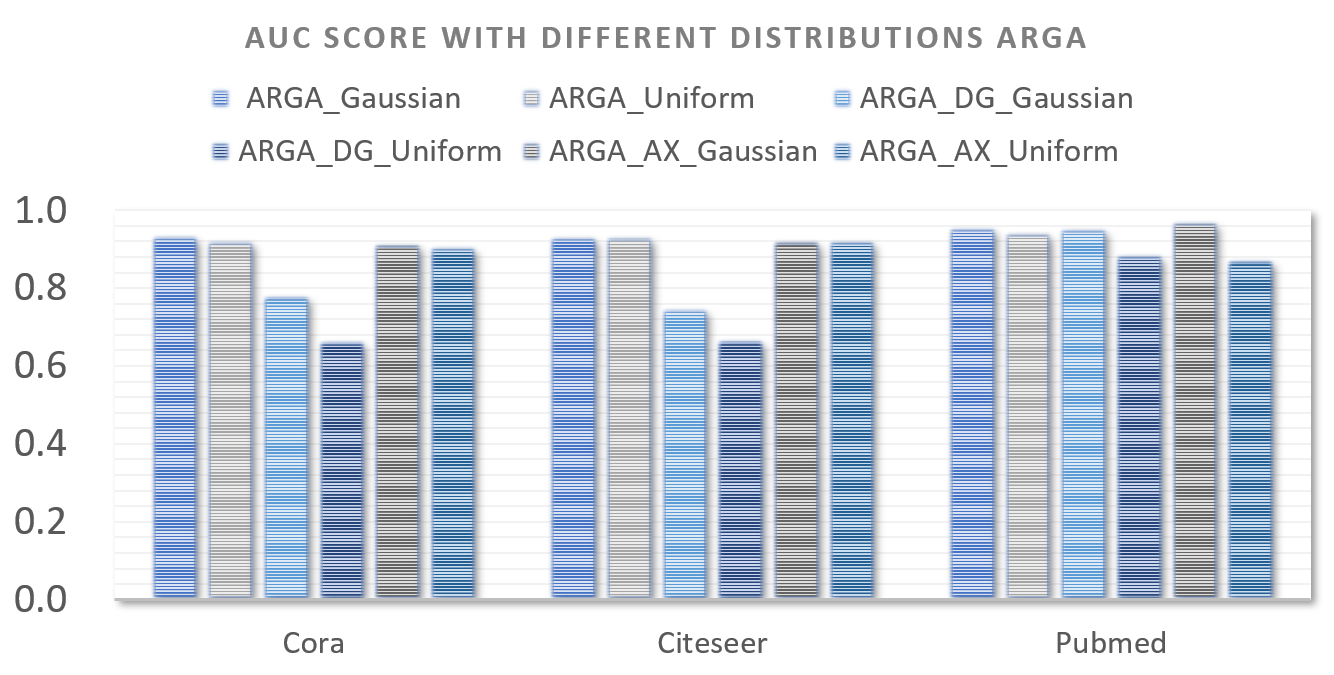}
		\includegraphics[width=0.49\linewidth,height=3.5cm]{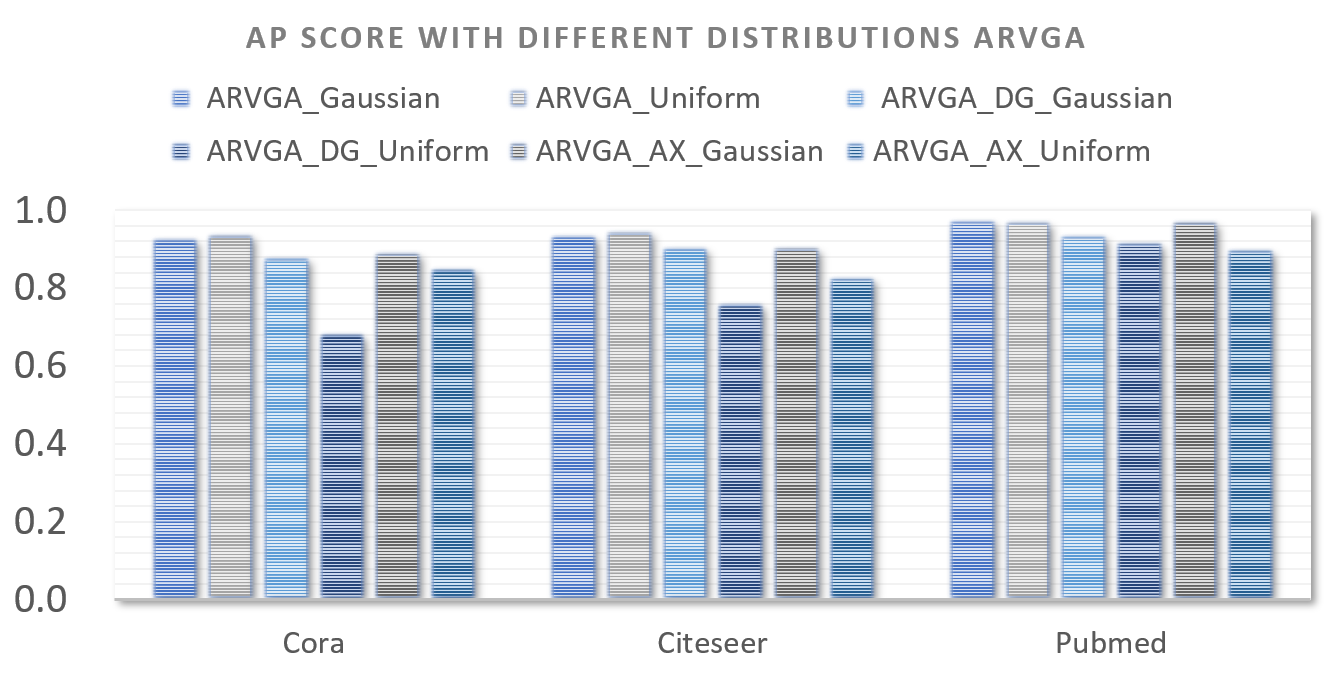}
		\includegraphics[width=0.49\linewidth,height=3.5cm]{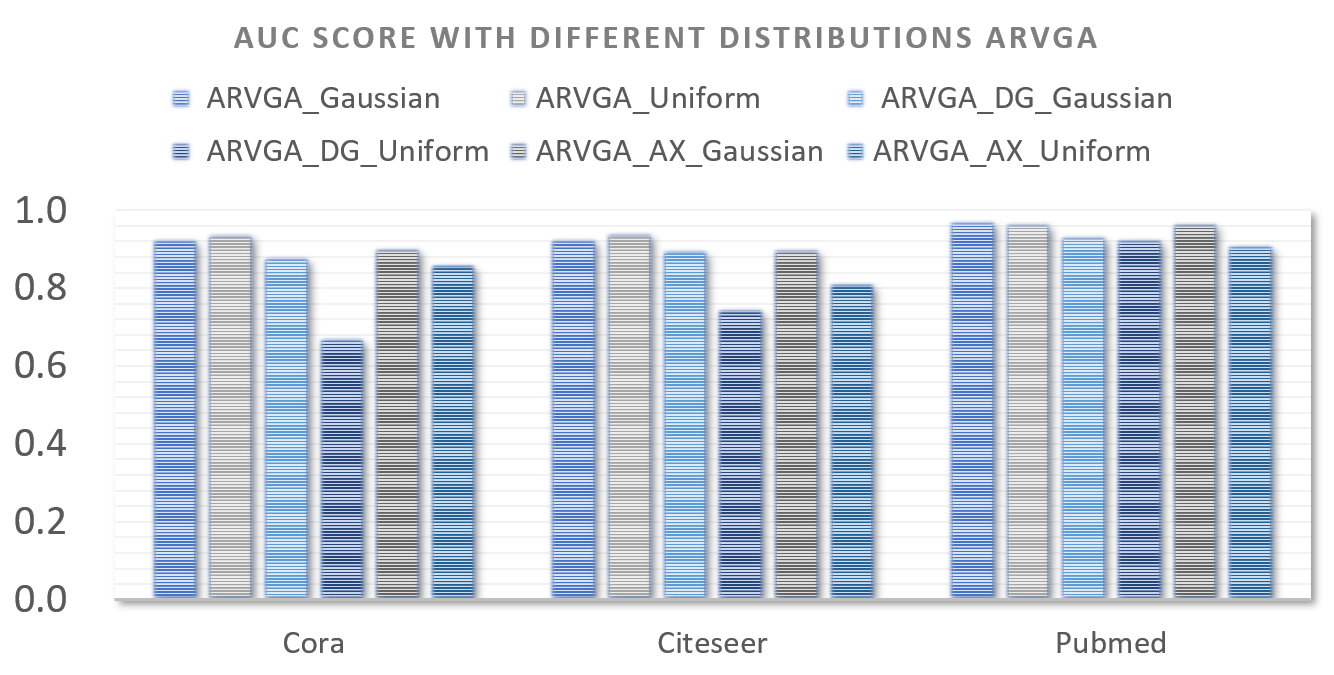}
		\caption{The ARGA related models comparison on the \textbf{link prediction} task with different prior distributions.
		} \label{fig:link_prediction}
	\end{figure*}
	
	Furthermore, as we can see from the three tables, the clustering results from approaches BigClam and DeepWalk, which only consider one perspective information of the graph,  are inferior to the results from those which consider both topological information and node content of the graph. However, both purely GCNs-based approaches or the methods considering multi-view information still only obtain  sub-optimal results compared to the adversarially regularized graph convolutional models. 
	
	The wide margin in the results between ARGA and GAE (and the others) has further demonstrated the superiority of our adversarially regularized graph autoencoder.

	\vspace{.1cm}
	\noindent\textbf{Parameter Study. \ }
	We conducted experiments on Cora dataset with varying the dimension of embedding from 8 neurons to 1024 and report the results  in Fig \ref{fig:embedding_clustering}. 
    All metrics demonstrated a similar fluctuation as the dimension of the embedding is increased. We cannot extract apparent trends to represent the relations between the embedding dimensions and the score of each clustering metric. This observation indicates that the unsupervised clustering task is more sensitive to the parameters compared to the supervised learning tasks (e.g., link prediction in Section \ref{sec:link}). 
	
	
	
	
		\vspace{.1cm}
	\noindent\textbf{Graph Visualization with Linkages. \ }
	\begin{figure}[htpb]
		\centering
		\includegraphics[ width=0.49\linewidth]{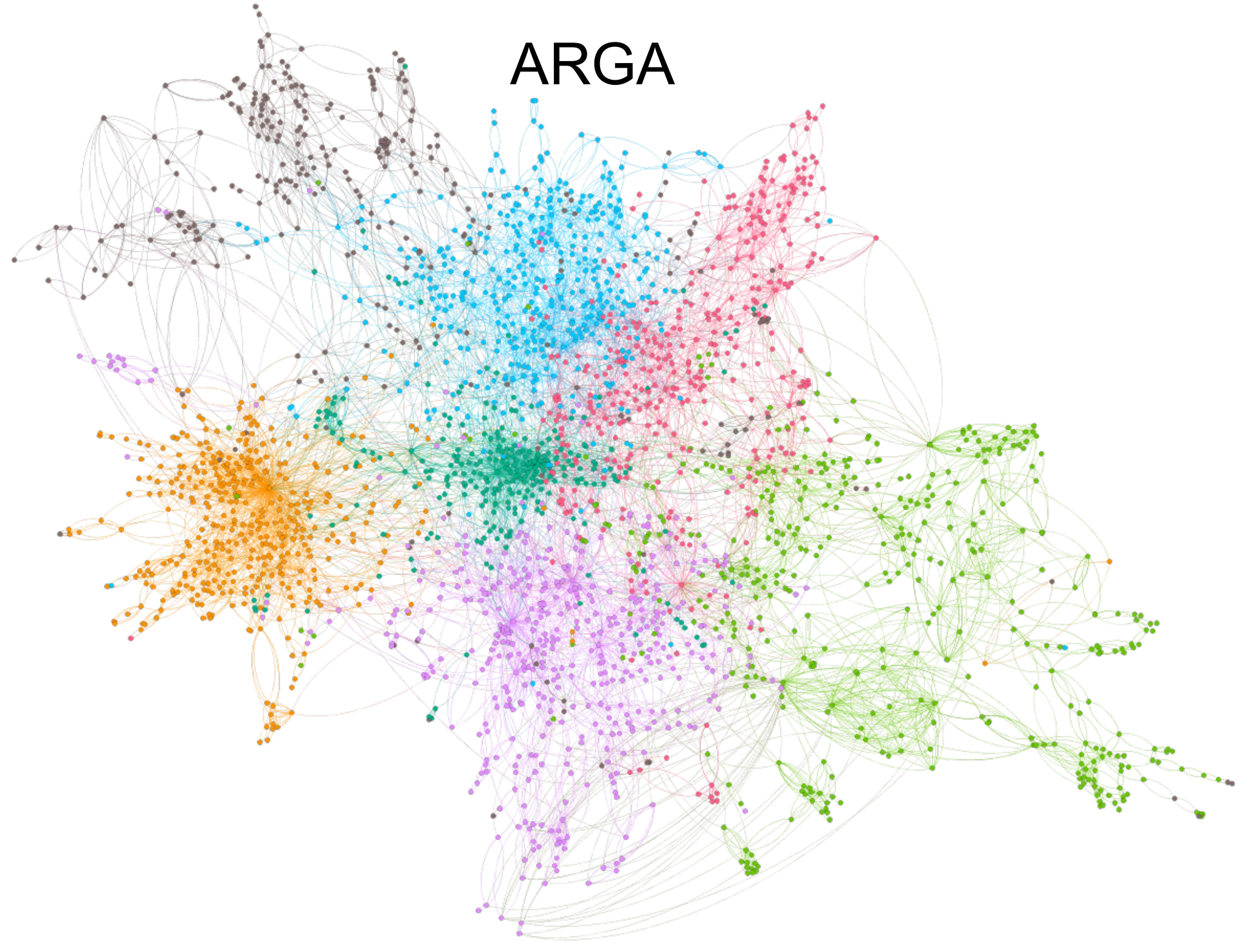}
		\includegraphics[ width=0.49\linewidth]{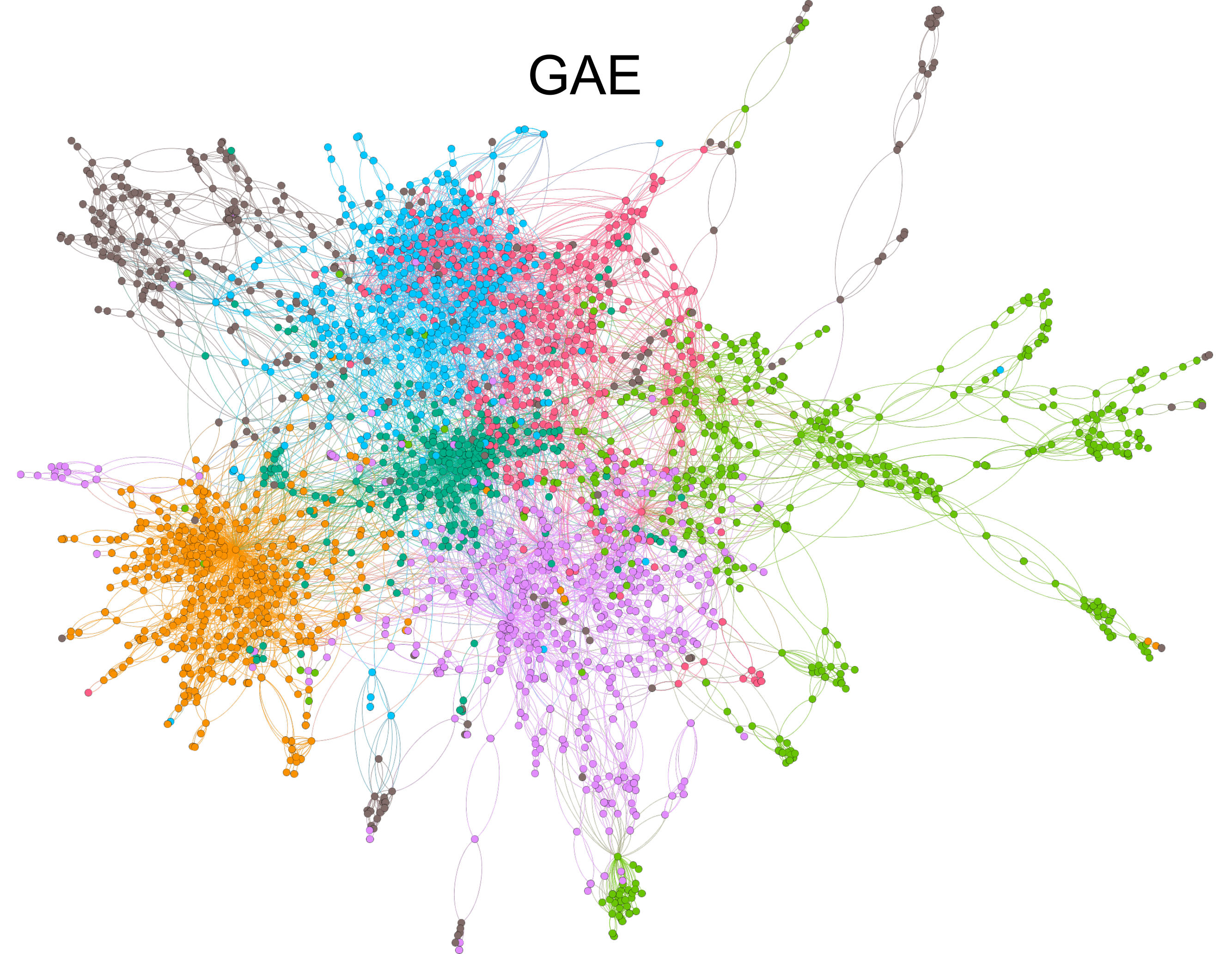}
		\caption{Visualization with edges of the latent space of unsupervised ARGA and GAE trained on Cora data set. Colors indicate different clusters, and edges are represented with the links between nodes. Best view for both models} \label{fig:embedding_visualization}
	\end{figure}
	
    Inspired by \cite{kipf2016variational}, we visualized the well-learned latent space with the linkages of both GAE and our proposed ARGA trained on Cora data set. As shown in Fig. \ref{fig:embedding_visualization}, many nodes in the latent space of GAE (Right side) which belong to the GREEN cluster have been located nearer to the PINK cluster. Similar circumstance happened in the bond between the RED cluster and the BLUE cluster, where some of nodes of RED mixed in the BLUE cluster. This could be caused by the unregularized embedding space, which is free for any structure. Adversarially regularized embedding shows better visualization with clear boundary line between two clusters. Considering the only difference between ARGA and the GAE is the adversarial training regularization scheme, it is reasonable to claim that adversarial regularization is helpful to enhance the quality of graph embedding.

	\subsection{ARGA Architectures Comparison} \label{sec:comparison}
	In this section, we  construct six versions of the model: \textit{adversarially regularized graph autoencoder} (ARGA), \textit{adversarially regularized graph autoencoder with graph convolutional decoder} (ARGA\_DG) and \textit{adversarially regularized graph autoencoder for reconstructing both graph structure and node content} (ARGA\_AX) and their variational versions. Meanwhile, we conduct all experiments with a prior Gaussian distribution and a prior Uniform Distribution respectively for every model. We analyze the comparison experiments and try to figure out the reasons behind the results. The experimental results are illustrated in Fig, \ref{fig:clustering} and \ref{fig:link_prediction}.
	
	\vspace{.1cm}
	\noindent\textbf{Gaussian Distribution vs Uniform Distribution. \ }
    The performance of the proposed models is not very sensitive to the prior distributions, especially for the node clustering task. As shown in Fig. \ref{fig:clustering}, if we compare the results of two distributions with the same metric, the results from one same model, in most cases, are very similar. 
	
    As for the link prediction (Fig. \ref{fig:link_prediction}), the Uniform distribution dramatically lowers the performance of ARGA\_DG on all datasets and metrics, compared to the results with Gaussian distribution. ARGA and its variational version are not as sensitive to the different distributions as ARGA\_DG models. The standard version of ARGA with Gaussian distribution slightly outperforms the ones with Uniform distribution. The situation reversed with the variational ARGA models. 
	
	\noindent\textbf{Decoders and Reconstructions. \ }
    As shown in Fig \ref{fig:link_prediction}, the ARGA with the Gaussian distribution and inner product decoder for reconstructing graph structure has a significant advantage in link prediction since $p(\hat{\mathbf{A}}_{ij} = 1 |\mathbf{z}_i, \mathbf{z}_j)$ is designed to predict whether there is a link between two nodes. Simply replacing the decoder with graph convolutional layers to reconstruct adjacency matrix $\hat{\mathbf{A}}$ (ARGA\_DG) has a sub-optimal performance in link prediction compared to ARGA. According to the statistic in Fig. \ref{fig:clustering}, although the performance of ARGA\_DG on clustering is comparable with original ARGA, there is still a gap between these two variations. Two graph convolutional layers in the decoder cannot effectively decode the topological information of the graph, which leads to the sub-optimal results. The model with graph convolutional decoder for reconstructing both topological information $\mathbf{A}$ and node content $\mathbf{X}$ (ARGA\_AX) may prove this hypothesis. As can be seen in Fig.  \ref{fig:clustering} and \ref{fig:link_prediction}, ARGA\_AX has dramatically improved the performance on both link prediction and clustering compared to ARGA\_DG which purely reconstructs the topological structure. ARGA and ARGA\_AX have very similar performances on both link prediction and clustering. The variational version of ARGA\_AX (ARVGA\_AX) has outstanding performance on clustering which has achieved 12.2\% improvement on clustering accuracy on Cora dataset and 5.4\% improvement on Citeseer dataset compared to ARVGA. 
	
	
	\subsection{Time Complexity on Convollution} \label{sec:Time}
	Our graph encoder requires the computation $\mathbf Z^{'} = \phi(\widetilde{\mathbf{D}}^{-\frac{1}{2}}\widetilde{\mathbf{A}}\widetilde{\mathbf{D}}^{-\frac{1}{2}}\mathbf{Z}^{(l)}\mathbf{W}^{(l)})$, which can be computed efficiently using sparse matrix computation. Specifically, let $\mathbf P=\widetilde{\mathbf{D}}^{-\frac{1}{2}}\widetilde{\mathbf{A}}\widetilde{\mathbf{D}}^{-\frac{1}{2}}$  , which is the Laplacian matrix. As $\widetilde{\mathbf{D}}$ is a diagonal matrix, the inverse of $\widetilde{\mathbf{D}}$ is the inverse of its diagonal values with time complexity $O(|V|)$. Let $\mathbf{W}^{(l)} \in \mathbbm{R}^{m\times d}$, and $\mathbf{Z}^{(l)} \in \mathbbm{R}^{n\times m}$. The  complexity of our convolution operation is $O(|\mathbf E|md)$, as $\widetilde{\mathbf{A}}\mathbf{Z}^{(l)}$ can be efficiently implemented as a product of a sparse matrix with a dense matrix (See \cite{kipf2016variational} for details). 
	
    We conducted experiments with six ARGA models and two GAE models for the training time comparison. We conducted 200 training epochs for link prediction task of each model on Cora data set and report the average time for the comparison. The results are shown in Fig. 9. The results show that ARGA models take more time than original GAE models due to the additional regularization module in the architecture. ARGA\_AX model requires more computation for simultaneously reconstructing both topological structure (A) and node characteristics (X). 
	
		\begin{figure}[htpb]
		\centering
		\includegraphics[ width=\linewidth]{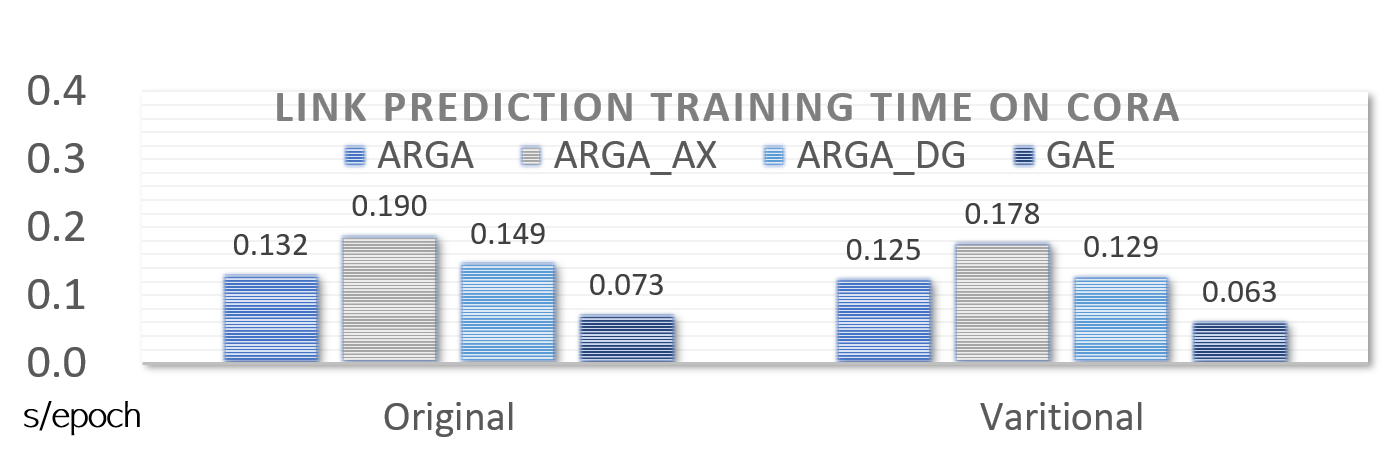}
		\caption{Average training time per epoch of link prediction. (Left) Original Architectures; (Right) Varitional Architectures.} \label{fig:time}
	\end{figure}
	
	\section{Conclusion and Future Work} \label{sec:conclusion}

        In this paper, we proposed a novel adversarial graph embedding framework for graph data. We argue that most existing graph embedding algorithms are unregularized methods that ignore the data distributions of the latent representation and suffer from inferior embedding in real-world graph data. We proposed an adversarial training scheme to \textit{regularize} the latent codes and enforce the latent codes to match a prior distribution. The adversarial module is jointly learned with a graph convolutional autoencoder to produce a robust representation.  We also exploited some interesting variations of ARGA like ARGA\_DG and ARGA\_AX to discuss the impact of graph convolutional decoder for reconstructing both graph structure and node content. Experiment results demonstrated that our algorithms ARGA and ARVGA outperform baselines in link prediction and node clustering tasks.	
	
	There are several directions for the adversarially regularized graph autoencoders (ARGA). We will investigate how to use the ARGA model to generate some realistic graphs \cite{you2018graphrnn}, which may help discover new drugs in biological domains. We will also study how to incorporate label information into ARGA to learn robust graph embedding. 

\section*{Acknowledgment}

This research was funded by the Australian Government through the Australian Research Council (ARC) under grants 1) LP160100630 partnership with Australia Government Department of Health and 2) LP150100671 partnership with Australia Research Alliance for Children and Youth (ARACY) and Global Business College Australia (GBCA). 
We  acknowledge the support of NVIDIA Corporation and MakeMagic Australia with the donation of GPU used for this research.

\bibliographystyle{IEEEtran}
\bibliography{IEEEabrv,tcyb-template}

\balance

\end{document}